%% file: main.tex
\documentclass{article}

\PassOptionsToPackage{numbers, compress}{natbib}

\usepackage[preprint]{neurips_2025}

\usepackage[utf8]{inputenc} 
\usepackage[T1]{fontenc}    
\usepackage{hyperref}       
\usepackage{url}            
\usepackage{booktabs}       
\usepackage{amsfonts}       
\usepackage{nicefrac}       
\usepackage{microtype}      
\usepackage{xcolor}         
\usepackage{textcomp}
\usepackage{graphicx}
\usepackage{multirow}
\usepackage{color}
\usepackage{threeparttable} 
\usepackage{tablefootnote} 
\usepackage{arydshln} 
\usepackage{amssymb} 
\usepackage{pifont} 
\usepackage{wrapfig}
\usepackage{verbatim}
\usepackage{enumitem}
\usepackage{amsmath}
\usepackage{colortbl}
\usepackage{setspace}
\usepackage{hhline}
\usepackage{subcaption}
\usepackage{makecell}
\usepackage{mathrsfs}
\usepackage{tikz}
\usepackage{tabularx}
\usepackage{fontawesome5}
\usepackage{algorithm}
\usepackage{algorithmic}

\newcommand{\MYMETHOD}{\textsc{MindLoom}}

\newcolumntype{C}[1]{>{\centering\arraybackslash}p{#1}}

\title{\MYMETHOD: Composing Thought Modes for Frontier-Level Reasoning Data Synthesis}

\author{%
  \vspace{-25pt}\\
  \textbf{\small Haiyang Shen$^{1,*}$,\quad Taian Guo$^{1,*}$,\quad Xuanzhong Chen$^{2,*}$,\quad Mugeng Liu$^{1}$,\quad Weichen Bi$^{1}$}\\
  \textbf{\small Wenchun Jing$^{1}$,\quad Sixiong Xie$^{1}$,\quad Zhuofan Shi$^{1}$,\quad Yudong Han$^{1}$,\quad Chongyang Pan$^{1}$}\\
  \textbf{\small Siqi Zhong$^{1}$,\quad Jinsheng Huang$^{1}$,\quad Ming Zhang$^{1,\dag}$,\quad Yun Ma$^{1,\dag}$}\vspace{8pt}\\
  $^{1}$Peking University \quad $^{2}$Tsinghua University \\
  \texttt{\small hyshen@stu.pku.edu.cn}, \texttt{\small taianguo@stu.pku.edu.cn}, \texttt{\small cxz23@mails.tsinghua.edu.cn},\\
  \texttt{\small mzhang\_cs@pku.edu.cn}, \texttt{\small mayun@pku.edu.cn}\vspace{8pt}\\
  Explore codes and datasets at:~\, \url{https://github.com/EachSheep/MindLoom}\\
  \vspace{3em}
  $^*$Equal contribution. $^\dag$Corresponding author.
  \vspace{-48pt} \\
}

\begin{document}

\maketitle

\input{sections/0_Abstract}

\input{sections/1_Introduction}
\input{sections/2_RelatedWork}
\input{sections/3_Method}
\input{sections/4_Evaluation}
\input{sections/5_Conclusion}

\bibliographystyle{plainnat}
\bibliography{reference}

\input{sections/Appendix}

\end{document}

%% file: sections/0_Abstract.tex
\begin{abstract}

Although LLMs have made substantial progress in reasoning, systematically producing frontier-level reasoning data remains difficult. Existing synthesis methods often have limited visibility into the structural factors that govern problem difficulty, which can result in narrow diversity and unstable difficulty control. In this work, we view the difficulty of a reasoning problem as arising from the accumulation of atomic knowledge-reasoning transformations, which we term \textit{thought modes}. Building on this perspective, we propose \MYMETHOD, a framework for synthesizing frontier-level reasoning data through compositional thought mode engineering. Given a collection of hard problems with verified solutions, \MYMETHOD~first decomposes those solutions into thought mode chains that reveal each problem's construction logic. It then trains a retrieval model that matches problem states to compatible thought modes, providing guidance on which reasoning challenges to introduce during synthesis. New problems are composed by iteratively applying retrieved thought modes to seed questions, with distribution-aligned sampling to encourage diverse reasoning coverage. Finally, a rollout-based judging stage labels generated questions by difficulty and supplies judged-correct responses for supervised fine-tuning. We evaluate \MYMETHOD~on nine benchmarks covering five STEM disciplines and four mathematical reasoning tasks across multiple model families and sizes. Models fine-tuned on \MYMETHOD-generated data achieves favorable performances over base models, distillation, and external-data baselines across the reported benchmarks. Ablation studies indicate the contribution of each component, and further analysis suggests that \MYMETHOD~covers a broad range of reasoning patterns while maintaining useful difficulty control. We have open-sourced our implementation at \url{https://github.com/EachSheep/MindLoom}.

\end{abstract}

%% file: sections/1_Introduction.tex
\section{Introduction}
\label{sec:Introduction}

LLMs have made substantial progress on complex reasoning tasks across mathematics, science, and other domains~\citep{openai2024gpt4technicalreport,Guo_2025,yang2025qwen3technicalreport}. As models advance, evaluation increasingly relies on benchmarks that remain challenging at the frontier of model capability~\citep{phan2026hle,rein2023gpqagraduatelevelgoogleproofqa}. Yet manually constructing frontier-level reasoning problems is expensive, requires deep domain expertise, and is hard to scale.

Existing reasoning data construction falls into three categories. The first category focuses on \textit{reasoning data synthesis}, where strong models generate questions or curate solution traces~\citep{Guo_2025,luo2025wizardmathempoweringmathematicalreasoning,zhan2025mathsmithextremelyhardmathematical}. The second category targets \textit{evaluation benchmark construction}, in which human experts craft questions that push the boundary of model capability~\citep{phan2026hle,rein2023gpqagraduatelevelgoogleproofqa,balunovic2025matharena}. The third category employs \textit{data selection strategies} to curate high-value training data from large pools~\citep{zhang2026expandingreasoningpotentialfoundation,pmlr-v235-xia24c,gu2025dataselectionoptimalcontrol}.

Despite their contributions, these approaches face notable limitations. Synthesis methods rarely model what makes a problem structurally hard, so generated questions tend to be superficially varied but homogeneous in their reasoning composition. Benchmark construction by experts produces high-quality questions but is slow, costly, and hard to scale to training-data volumes. Data selection identifies valuable samples but cannot create new problems, and its coverage is bounded by source-pool diversity. Across the three categories, controllable frameworks for the compositional structure of reasoning difficulty remain underdeveloped.

To address these limitations, we first ask: what makes a reasoning problem hard? Rather than treating difficulty as a monolithic property, we model it as a composite of atomic reasoning requirements. Each such requirement introduces specific knowledge and demands a corresponding derivation step. We call these atomic requirements \textit{thought modes}. This abstraction turns difficulty control into a compositional operation: new problems can be constructed by selecting and combining reusable thought modes in novel configurations. Based on this perspective, we propose \MYMETHOD, which synthesizes high-quality reasoning training data through compositional thought mode engineering.

\MYMETHOD~addresses four challenges through a four-stage pipeline. First, to learn how existing hard problems are constructed, we analyze verified solutions in reverse, which we term \textit{reverse engineering}, producing a chain of thought modes that serves as each problem's construction blueprint (Section~\ref{sec:extraction}). Second, to decide which thought mode applies to a given problem state, we train an embedding-based retrieval model on the extracted chains (Section~\ref{sec:retrieval}). Third, to compose new problems with broad reasoning coverage, we iteratively apply retrieved thought modes under a distribution-aligned sampling scheme that discourages concentration on common reasoning types (Section~\ref{sec:synthesis}). Fourth, to assemble training data, we run multi-rollout inference with LLM-based judging, filter by source provenance, and convert judged-correct rollouts into supervised fine-tuning records (Section~\ref{sec:rollout_filtering}).

We evaluate \MYMETHOD~on nine benchmarks spanning five STEM disciplines and four mathematical reasoning tasks, including competition-level problems. Across the Qwen3 and Qwen3.5 families, models fine-tuned on \MYMETHOD-generated data are compared against direct base-model inference, knowledge distillation (DS-V3.2 Distill), and external curated reasoning corpora (MegaScience, OpenThought). Results show consistent improvements across benchmarks, with especially large gains on competition-style mathematical reasoning. Ablation studies, distribution analysis, and case studies further verify the contribution of each pipeline stage.

Our main contributions are as follows:
\begin{itemize}[leftmargin=*, topsep=2pt, itemsep=2pt, parsep=0pt, partopsep=0pt]
\item We propose the concept of \textit{thought modes}, defined as atomic knowledge-reasoning transformations whose composition gives rise to problem difficulty, and develop a reverse-engineering procedure that decomposes expert-level problems into thought mode chains, offering a structural perspective on reasoning difficulty.
\item We present \MYMETHOD, a framework that integrates thought mode retrieval, distribution-aligned compositional synthesis, rollout-based judging, and source-aware conversion into a single pipeline for reasoning data construction.
\item We validate \MYMETHOD~on nine benchmarks across four model sizes from two model families. Models trained on \MYMETHOD-generated data achieve favorable performances over various types of baselines, supporting the effectiveness of compositional thought mode engineering.
\end{itemize}

%% file: sections/2_RelatedWork.tex
\section{Related Work}
\label{sec:RelatedWork}

This section reviews related work from three perspectives: reasoning data synthesis (Section~\ref{sec:rw_synthesis}), evaluation benchmark construction (Section~\ref{sec:rw_benchmark}), and data selection for reasoning (Section~\ref{sec:rw_selection}).

\subsection{Reasoning Data Synthesis}
\label{sec:rw_synthesis}

Reasoning data synthesis methods can be grouped into three lines. \textit{Knowledge distillation and open data-recipe} methods leverage strong reasoning models to generate or curate solution traces for existing problems, transferring frontier-model reasoning capability to smaller ones at scale~\citep{Guo_2025,openr1_math_220k,guha2026openthoughts}. \textit{Direct generation and augmentation} methods prompt LLMs to create new questions through templates, evolutionary prompting, or surface-level transformations~\citep{luo2025wizardmathempoweringmathematicalreasoning,yu2024metamath,xu2024wizardlmempoweringlargepretrained}; the diversity and difficulty of the output depend largely on prompt design and seldom expose the structural factors that make problems hard. \textit{Structured and skill-compositional} methods build symbolic generators or learn policies that assemble difficult problems from extracted skills~\citep{liu2025synlogicsynthesizingverifiablereasoning,shah2024aiassistedgenerationdifficultmath,zhan2025mathsmithextremelyhardmathematical}; they rely on logic templates or reinforcement learning and typically operate within a single domain such as mathematics.

\MYMETHOD~differs in that it explicitly models how hard problems are constructed. Rather than treating difficulty escalation as opaque prompting or domain-specific policy learning, \MYMETHOD~reverse-engineers verified solutions across broad STEM domains into atomic thought mode transformations and composes new problems through learned retrieval and distribution-aligned sampling, without RL training or symbolic templates.

\subsection{Evaluation Benchmark Construction}
\label{sec:rw_benchmark}

Evaluation benchmarks for reasoning have evolved from standardized tests to frontier-level challenges. Early benchmarks such as GSM8K and MATH focus on grade-school and competition-level mathematics with well-defined difficulty tiers~\citep{cobbe2021trainingverifierssolvemath,hendrycks2021measuringmathematicalproblemsolving}, but are now largely saturated by frontier models. Recent efforts push toward the boundary of human and model capability, including HLE, GPQA, and FrontierMath~\citep{phan2026hle,rein2023gpqagraduatelevelgoogleproofqa,glazer2025frontiermathbenchmarkevaluatingadvanced}, with complementary work on broad expert domains and uncontaminated olympiad-level evaluation~\citep{pteam2025supergpqascalingllmevaluation,balunovic2025matharena}. Such benchmarks are typically curated by domain experts, which produces high-quality questions but is slow, costly, and hard to scale. A separate line of work explores dynamic or synthetic benchmark construction by extracting and perturbing reasoning structures~\citep{NEURIPS2024_f5198bc2,pmlr-v267-xu25n}, highlighting the value of controllable reasoning structures for evaluation.

\MYMETHOD~draws on the construction principles that make expert-crafted questions hard, that compositional reasoning requirements accumulate across multiple steps, but applies them to training-data synthesis. By reverse-engineering the construction logic of solved hard problems and composing thought modes in novel configurations, \MYMETHOD~produces new training problems whose construction logic mirrors patterns observed in expert-crafted ones, with greater diversity and scalability.

\subsection{Data Selection for Reasoning}
\label{sec:rw_selection}

Recent work on data selection for reasoning training aims to identify the most valuable samples from large data pools. \textit{Quality-based} methods use signals such as gradient influence, model uncertainty, learned quality ratings, and training-dynamics objectives to prioritize samples that improve downstream capabilities~\citep{pmlr-v235-xia24c,NEURIPS2024_b130a569,pmlr-v235-wettig24a}. \textit{Pattern-based} methods abstract structural patterns from reasoning traces and select data that covers diverse, high-value patterns; for example, CoTP extracts atomic reasoning patterns from chain-of-thought sequences~\citep{zhang2026expandingreasoningpotentialfoundation,yang2025select2reasonefficientinstructiontuningdata}. The shared idea is that a carefully chosen subset can match or exceed training on the full pool, improving training efficiency.

\MYMETHOD~complements data selection by creating new reasoning data through compositional synthesis rather than selecting from an existing pool. Among existing work, CoTP~\citep{zhang2026expandingreasoningpotentialfoundation} is most related in representation design, also decomposing reasoning into atomic patterns, but it applies the patterns to select examples from an existing pool. \MYMETHOD~instead uses extracted thought modes to generate new problems, so its coverage is less directly bounded by the diversity of the source pool.

%% file: sections/3_Method.tex
\section{Method}
\label{sec:Method}

\subsection{Overview}
\label{sec:overview}

\begin{figure}[ht!]
    \centering
    \captionsetup{skip=2pt}
    \includegraphics[width=\linewidth]{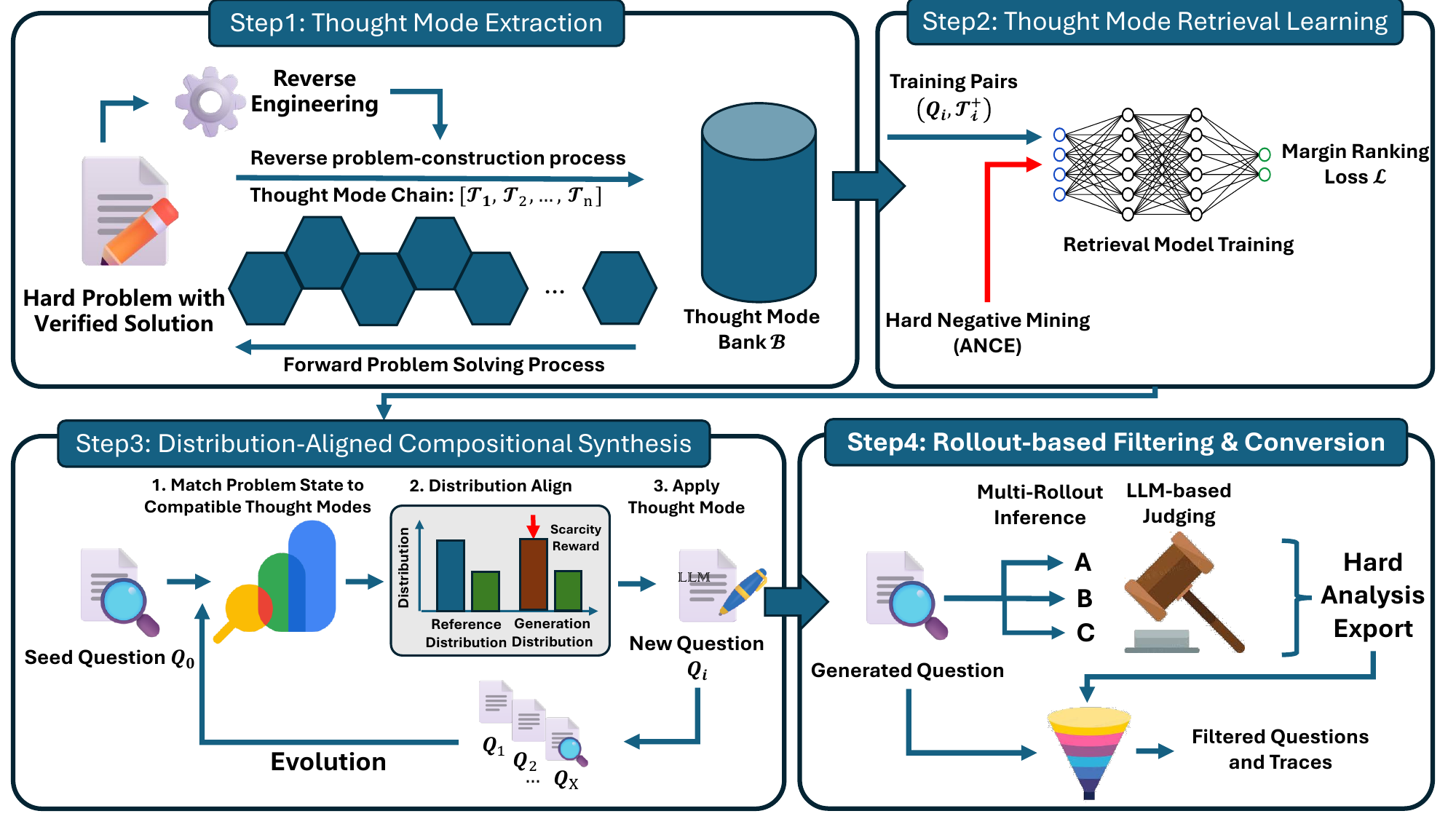}
    \caption{Overview of the \MYMETHOD~pipeline. \textbf{Step~1}: hard problems with verified solutions are reverse-engineered into thought mode chains, populating a thought mode bank $\mathcal{B}$. \textbf{Step~2}: training pairs $(Q_{i-1}, \mathcal{T}_i^+)$ and mined hard negatives train a retrieval model via margin ranking loss $\mathcal{L}$. \textbf{Step~3}: from a seed question $Q_0$, the pipeline iteratively matches compatible thought modes, applies distribution-aligned scoring with a scarcity reward, and integrates the selected mode via an LLM to yield evolved questions $Q_1, Q_2, \ldots$ \textbf{Step~4}: generated questions undergo multi-rollout inference and LLM judging, converting judged-correct traces into SFT records.}
    \label{fig:overview}
    \vspace{-0.9em}
\end{figure}

Figure~\ref{fig:overview} illustrates the four-stage \MYMETHOD~pipeline. Step~1 (\textit{Thought Mode Extraction}, Section~\ref{sec:extraction}) addresses \textit{what makes problems hard}: it reverse-engineers verified solutions into thought mode chains, populating a thought mode bank $\mathcal{B}$. Step~2 (\textit{Retrieval Learning}, Section~\ref{sec:retrieval}) addresses \textit{which thought mode to apply next}: it trains an embedding-based retriever on these extracted chains. Step~3 (\textit{Distribution-Aligned Synthesis}, Section~\ref{sec:synthesis}) addresses \textit{how to generate diverse problems}: it iteratively retrieves and selects compatible thought modes for seed questions via a similarity-scarcity trade-off, applying them via an LLM to yield increasingly complex questions. Step~4 (\textit{Rollout-Based Filtering}, Section~\ref{sec:rollout_filtering}) addresses \textit{how to prepare training data}: it performs multi-rollout judging and provenance filtering, converting correct rollouts into SFT records.

\subsection{Thought Mode Extraction}
\label{sec:extraction}

To synthesize hard problems, we first need to understand what makes existing problems hard. Rather than treating difficulty as an opaque property, \MYMETHOD~decomposes it into atomic components by analyzing how hard problems could have been constructed. This decomposition is not meant to recover the original authoring history; it is an operational inverse curriculum that yields progressively simpler sub-problems and is useful for synthesis regardless of whether it matches the original construction intent. We first define the core abstractions and then describe the extraction process. Detailed prompt templates and extended examples are provided in Appendix~\ref{app:thought_mode_format} and~\ref{app:reverse_engineering}.

\textbf{Thought Mode.}
A thought mode $\mathcal{T} = (S_\mathrm{sum}, S_\mathrm{det}, K_\mathrm{gen}, K_\mathrm{spec})$ is an atomic knowledge-reasoning transformation that adds one layer of complexity to a problem. The four components capture complementary aspects: $S_\mathrm{sum}$ provides a high-level summary of the transformation type (e.g., ``introduce integral computation''); $S_\mathrm{det}$ details the specific modification; $K_\mathrm{gen}$ captures the general, transferable knowledge required (e.g., theorems, formulas, or principles); and $K_\mathrm{spec}$ records the problem-specific parameters involved (e.g., constraints, numerical values, or domain setup). Each thought mode represents a single, well-defined reasoning requirement that a solver must handle.

\textbf{Thought Mode Chain.}
A thought mode chain $\mathcal{C} = [\mathcal{T}_1, \mathcal{T}_2, \ldots, \mathcal{T}_n]$ is an ordered sequence of thought modes that encodes the construction blueprint of a problem. When applied sequentially to a seed question $Q_0$, the chain produces increasingly complex questions $[Q_1, Q_2, \ldots, Q_n]$, where the final question $Q_n$ approximates the original hard problem. The chain captures both the structural progression of difficulty and the specific knowledge-reasoning requirements at each layer.

\textbf{Reverse Engineering Process.}
We analyze the verified solution reversely, since reasoning dependencies are easier to isolate from the tail of a solution than from its head. Given a hard problem $Q$ with a verified solution $S = [s_1, s_2, \ldots, s_m]$, we partition $S$ into $k$ windows $[W_1, W_2, \ldots, W_k]$ from the tail, each containing a fixed number of consecutive steps. The extraction then proceeds in two phases:

\textit{Phase 1: Seed Generation.} Starting from the last window $W_k$, we generate a minimal seed question $Q_0$ whose solution requires only the steps in $W_k$. The principle of \textit{dependency isolation} ensures that any values computed in earlier steps are converted into explicit givens in $Q_0$. This makes the seed question fully self-contained and independently solvable.

\textit{Phase 2: Iterative Evolution.} Starting from $Q_0$, we iterate backwards through the remaining windows. At each step $i$, we perform \textit{dependency absorption}: the current question $Q_{i-1}$ is modified to remove an explicit given that corresponds to what window $W_{k-i}$ computes. This forces the solver to derive that value, effectively adding one thought mode of complexity. The process produces the evolved question $Q_i$ along with the thought mode $\mathcal{T}_i$ that describes the added reasoning requirement.

Formally, the process is:
\begin{align}
Q_0 &= \operatorname{Seed}(W_k, [s_1, \ldots, s_{m-|W_k|}]) \label{eq:seed} \\
Q_i, \mathcal{T}_i &= \operatorname{Absorb}(Q_{i-1}, W_{k-i}, [s_1, \ldots, s_{m - \textstyle\sum_{j=0}^{i}|W_{k-j}|}]), \quad i = 1, \ldots, k\!-\!1 \label{eq:absorb}
\end{align}
Both operations are implemented via LLM prompting with structured output (see Appendix~\ref{app:reverse_engineering} for details). Complete extraction of a single problem yields a thought mode chain $\mathcal{C} = [\mathcal{T}_1, \ldots, \mathcal{T}_{k-1}]$ and the corresponding question evolution sequence $[Q_0, Q_1, \ldots, Q_{k-1}]$. Applying this process to a corpus of hard problems produces a \textit{thought mode bank} $\mathcal{B}$ containing diverse atomic transformations.

\subsection{Thought Mode Retrieval Learning}
\label{sec:retrieval}

With a bank of extracted thought modes, the next challenge is determining which thought modes are applicable to a given problem state during synthesis. Not every thought mode is compatible with every problem: applying an algebraic transformation to a geometry problem, for example, may produce an incoherent result. To provide directional guidance for synthesis, we train an embedding-based retrieval model that matches problem states to compatible thought modes. The full training configuration is provided in Appendix~\ref{app:retrieval_details}.

\textbf{Training Data Construction.}
From the extracted thought mode chains, we construct training pairs $(Q_{i-1}, \mathcal{T}_i)$, where $Q_{i-1}$ is the problem state before applying thought mode $\mathcal{T}_i$. The query side is the question text of $Q_{i-1}$, and the document side is a serialized concatenation of $S_\mathrm{sum}, S_\mathrm{det}, K_\mathrm{gen}$, and $K_\mathrm{spec}$. A richer state representation is additionally embedded during synthesis. The exact serialization formats are provided in Appendix~\ref{app:retrieval_details}.

\textbf{Hard Negative Mining.}
We mine hard negatives by retrieving, under the current model, the nearest non-positive thought modes for each query, and refresh them every $R$ optimizer steps so that they remain challenging as the model improves. Random negatives are used only when the retrieved pool is too small. Index construction details are provided in Appendix~\ref{app:retrieval_details}.

\textbf{Training Objective.}
The retrieval model is trained with a margin ranking loss:
\begin{equation}
\mathcal{L} = \sum_{(Q_{i-1}, \mathcal{T}_i^+)} \sum_{\mathcal{T}_j^-} \max\!\left(0,\; \operatorname{sim}(Q_{i-1}, \mathcal{T}_j^-) - \operatorname{sim}(Q_{i-1}, \mathcal{T}_i^+) + \gamma\right)
\label{eq:retrieval_loss}
\end{equation}
where $\mathcal{T}_i^+$ is the correct thought mode, $\mathcal{T}_j^-$ are hard negatives, $\operatorname{sim}(\cdot, \cdot)$ is the embedding similarity, and $\gamma$ is the margin. This objective encourages the model to place compatible thought modes closer to the query problem state in the embedding space than incompatible ones.

\subsection{Distribution-Aligned Compositional Synthesis}
\label{sec:synthesis}

With the retrieval model providing directional guidance, we can now compose new problems by iteratively applying thought modes to seed questions. However, naive composition tends to concentrate on a small set of common thought mode types, which can lead to structurally homogeneous outputs. To mitigate this mode collapse and encourage broad coverage of the reasoning types present in the reference corpus, we design a three-sub-step synthesis loop with distribution-aligned sampling. Appendix~\ref{app:synthesis_details} gives the full algorithm.

\textbf{Sub-step 1: Match Problem State to Compatible Thought Modes.}
Starting from a seed question $Q_0$, \MYMETHOD~performs up to $n$ evolution steps. At each step $i$, the fine-tuned retrieval model (Section~\ref{sec:retrieval}) encodes the current problem state $Q_i$ and retrieves the top-$m$ most similar thought modes from the bank $\mathcal{B}$ as candidates for the next evolution.

\textbf{Sub-step 2: Distribution-Aligned Scoring.}
The retrieved candidates are then re-scored using a separate general-purpose embedding model, which provides the similarity term in a combined score that also accounts for cluster-level coverage. Specifically, we cluster the thought mode bank into $K$ clusters using K-Means~\citep{macqueen1967methods} on thought mode embeddings. Let $P_\mathrm{ref}$ denote the reference distribution of thought modes across clusters in the bank, and $P_\mathrm{gen}$ denote the empirical distribution of thought modes used in generation so far. For each candidate thought mode $\mathcal{T}_j$ belonging to cluster $c_j$, its selection score is:
\begin{equation}
\operatorname{score}(\mathcal{T}_j \mid Q_i) = \alpha \cdot \operatorname{sim}(Q_i, \mathcal{T}_j) + (1 - \alpha) \cdot \tanh\!\left(\log\!\left(1 + \frac{P_\mathrm{ref}(c_j)+\epsilon}{P_\mathrm{gen}(c_j)+\epsilon}\right)\right)
\label{eq:score}
\end{equation}
where $\alpha \in [0, 1]$ balances compatibility and coverage, and $\epsilon$ is a smoothing constant. The first term ensures the selected thought mode is semantically compatible with the current problem state. The second term is the bounded \textit{scarcity reward}: thought mode types that are underrepresented in the generated data relative to the reference distribution receive higher scores, while the $\tanh$ bound keeps the reward on a comparable scale to similarity. We align to $P_\mathrm{ref}$ rather than a uniform distribution to respect the natural frequency of reasoning types across domains.

The final selection is made through softmax sampling with temperature $\tau$:
\begin{equation}
P(\mathcal{T}_j \mid Q_i) = \frac{\exp\!\big(\operatorname{score}(\mathcal{T}_j \mid Q_i) / \tau\big)}{\sum_{l} \exp\!\big(\operatorname{score}(\mathcal{T}_l \mid Q_i) / \tau\big)}
\label{eq:softmax}
\end{equation}
This stochastic selection introduces controlled randomness while still favoring high-scoring candidates. The distribution tracker $P_\mathrm{gen}$ is updated after each synthesis step, enabling consistent diversity control throughout large-scale generation.

\textbf{Sub-step 3: Apply Thought Mode.}
The selected thought mode $\mathcal{T}_{i+1}$ is applied to the current problem state $Q_i$ through an LLM-based transformation to produce the evolved question $Q_{i+1}$. The LLM also returns an \texttt{is\_compatible} flag indicating whether the result is a coherent, solvable problem; incompatible-step handling follows Algorithm~1 in Appendix~\ref{app:synthesis_details}. The three sub-steps repeat for up to $n$ iterations, yielding a sequence of increasingly complex questions $Q_1, Q_2, \ldots, Q_n$.

\subsection{Rollout-Based Filtering and Conversion}
\label{sec:rollout_filtering}

Generated questions are not guaranteed to lie at the desired difficulty level or to admit a reliable training target, so we introduce a rollout-and-judge stage. For each generated question, we run three sampled solutions and use an LLM judge to label each rollout as correct or incorrect. The rollouts produce three difficulty categories (Appendix~\ref{app:rollout_filtering_details}): all-correct, partial, and all-wrong. For supervised fine-tuning (SFT) conversion, the criterion is the availability of a judged-correct target response: any generated question with at least one judged-correct rollout (whether all-correct or partial) is eligible. All-wrong questions are excluded because they lack a verified target.

We additionally filter generated data by source provenance before SFT conversion. The pipeline maps each generated seed back to its original question and datasource, and removes generated examples whose origins match the held-out benchmark sources listed in Appendix~\ref{app:corpus_details}. This separates source-aware benchmark filtering from difficulty labeling and keeps the training conversion aligned with the evaluation protocol.

%% file: sections/4_Evaluation.tex
\section{Experiments}
\label{sec:Evaluation}

\subsection{Experimental Setup}
\label{sec:setup}

\textbf{Benchmarks.}
We evaluate on nine benchmarks spanning five STEM disciplines and four mathematical reasoning tasks. Five benchmarks cover STEM reasoning: CS-Bench~\citep{song2025csbenchcomprehensivebenchmarklarge}, ChemBench~\citep{mirza2024largelanguagemodelssuperhuman}, HLE~\citep{phan2026hle}, MedQA~\citep{jin2020diseasedoespatienthave}, and SciBench~\citep{wang2024scibenchevaluatingcollegelevelscientific}. Four benchmarks target mathematical reasoning: MATH-500~\citep{hendrycks2021measuringmathematicalproblemsolving}, HMMT February 2025~\citep{balunovic2025matharena}, HMMT November 2025~\citep{balunovic2025matharena}, and AIME 2025~\citep{balunovic2025matharena}. We report pass@1 and pass@3, defined as the fraction of test items for which at least one correct answer is found among one or three independent rollouts, respectively. Throughout this paper, we use \textit{frontier-level} to refer to problems that current models cannot reliably solve under pass@3, rather than implying human-expert benchmark quality. Detailed benchmark descriptions are in Appendix~\ref{app:corpus_details}.

\textbf{Reference Corpus and Contamination Control.}
The thought-mode bank is built from 58,526 source problems aggregated from 16 datasets spanning broad STEM and expert-reasoning domains. Before SFT conversion, we trace each synthesized item back to its seed provenance and remove generated examples whose origin belongs to any held-out benchmark source. The full list of source datasets, their sizes, the provenance-filter configuration, and the domain composition of the final SFT set are provided in Appendix~\ref{app:corpus_details} and~\ref{app:data_composition}.

\textbf{Baselines.}
We compare against the following baselines. \textit{Base} directly evaluates the base model on the test sets without additional training. \textit{DS-V3.2 Distill} fine-tunes the base model on reasoning trajectories that DeepSeek V3.2~\citep{deepseekai2025deepseekv32pushingfrontieropen} generates for the original source problems without any structural decomposition, serving as a strong trajectory-level distillation baseline. \textit{MegaScience}~\citep{fan2025megasciencepushingfrontiersposttraining} is an external large-scale science reasoning dataset baseline. \textit{OpenThought}~\citep{guha2026openthoughts} is a reasoning dataset to train SOTA small reasoning models. \MYMETHOD~denotes our synthesized-data training recipe. All fine-tuned settings use the same foundation model and SFT configuration.

\textbf{Implementation Details.}
We evaluate on Qwen3~\citep{yang2025qwen3technicalreport} 4B and 8B, and Qwen3.5~\citep{qwen35blog} 4B and 9B. DeepSeek V3.2 serves as the chat model for thought mode extraction and synthesis. The retrieval model is initialized from Qwen3-Embedding-0.6B. Fine-tuning uses ms-swift~\citep{zhao2025swiftascalablelightweightinfrastructure} with full-parameter SFT for 3 epochs. Evaluation uses vLLM~\citep{kwon2023efficientmemorymanagementlarge} with three independent rollouts per test item and LLM-based judging. To isolate the effect of data quality from data quantity, all fine-tuned settings use exactly 9,230 SFT examples with identical SFT hyperparameters and evaluation protocol; the DS-V3.2 Distill set is randomly subsampled from a larger distillation pool to this same budget. Full hyperparameters are provided in Appendix~\ref{app:retrieval_details} and~\ref{app:synthesis_details}.

\subsection{Main Results}
\label{sec:main_results}

\begin{table}[t]
\centering
\caption{Main results. Each cell reports pass@1 / pass@3 (\%), with the best value in each model-size block in bold. HMMT-Feb.\ and HMMT-Nov.\ denote the February and November 2025 splits.}
\label{tab:main_results}
\tiny
\renewcommand{\arraystretch}{0.92}
\setlength{\tabcolsep}{2.6pt}
\resizebox{\textwidth}{!}{%
\begin{tabular}{rlccccccccc}
\toprule
\multirow{2}{*}{\textbf{Model}} & \multirow{2}{*}{\textbf{Setting}} & \multicolumn{5}{c}{\textbf{General and STEM}} & \multicolumn{4}{c}{\textbf{Mathematical Reasoning}} \\
\cmidrule(lr){3-7} \cmidrule(lr){8-11}
& & \textbf{CS-Bench} & \textbf{ChemBench} & \textbf{HLE} & \textbf{MedQA} & \textbf{SciBench} & \textbf{MATH-500} & \textbf{HMMT-Feb.} & \textbf{HMMT-Nov.} & \textbf{AIME 2025} \\
\midrule
\rowcolor{gray!10}
\multicolumn{11}{c}{\textit{Qwen3-2507 family (\%)}} \\
\midrule
4B & Base & 64.82 / 74.25 & 56.56 / 65.81 & 4.72 / 8.24 & 62.69 / 74.70 & 50.43 / 61.85 & 84.40 / 90.20 & 20.00 / 26.67 & 26.67 / 30.00 & 30.00 / 33.33 \\
4B & DS-V3.2 Distill & 63.62 / 78.87 & 52.16 / 69.79 & 5.16 / 11.40 & 60.17 / 76.28 & 47.90 / 64.90 & 81.20 / 91.60 & 23.33 / 33.33 & 23.33 / 30.00 & 26.66 / 36.70 \\
4B & MegaScience & 64.53 / 77.06 & 57.32 / 67.10 & 5.28 / 11.16 & 60.48 / 77.92 & 50.14 / 67.92 & 83.80 / 91.40 & 33.33 / 40.00 & 26.67 / 33.33 & 30.00 / 50.00 \\
4B & OpenThought & 65.85 / 78.75 & 60.70 / 72.26 & 5.72 / 11.32 & 59.78 / 74.94 & 49.42 / 69.51 & \textbf{85.20} / 94.00 & \textbf{40.00} / \textbf{46.67} & 30.00 / \textbf{40.00} & \textbf{43.33} / \textbf{56.67} \\
4B & \MYMETHOD & \textbf{68.29} / \textbf{80.16} & \textbf{61.72} / \textbf{74.78} & \textbf{5.96} / \textbf{12.12} & \textbf{65.12} / \textbf{81.30} & \textbf{53.61} / \textbf{71.53} & \textbf{85.20} / \textbf{98.60} & 33.33 / 43.33 & \textbf{33.33} / \textbf{40.00} & 40.00 / 53.33 \\
\midrule
8B & Base & 62.79 / 73.54 & 54.05 / 64.63 & 4.76 / 8.48 & 63.00 / 78.47 & 52.60 / 65.03 & 82.40 / 91.00 & 30.00 / 33.33 & 30.00 / 33.33 & 36.67 / 46.67 \\
8B & DS-V3.2 Distill & 65.97 / \textbf{81.52} & 53.07 / 69.16 & 5.04 / 10.36 & 67.47 / 81.78 & 52.31 / 68.93 & 84.40 / 93.00 & 14.81 / 33.33 & 30.00 / 43.33 & 36.67 / 40.00 \\
8B & MegaScience & 66.62 / 79.37 & 54.72 / 70.41 & 5.28 / 10.76 & 67.32 / 82.95 & 54.05 / 69.65 & 85.20 / 93.20 & 36.67 / 40.00 & \textbf{46.67} / \textbf{53.33} & 43.33 / 46.67 \\
8B & OpenThought & 67.50 / 80.57 & 55.97 / 70.10 & 6.08 / 11.36 & 66.92 / 81.07 & 56.21 / 68.93 & \textbf{87.80} / 97.20 & \textbf{46.67} / 50.00 & 43.33 / 50.00 & 40.00 / 53.33 \\
8B & \MYMETHOD & \textbf{69.65} / 81.40 & \textbf{59.04} / \textbf{72.97} & \textbf{6.64} / \textbf{11.64} & \textbf{70.77} / \textbf{86.80} & \textbf{57.08} / \textbf{73.41} & 86.80 / \textbf{97.40} & 43.33 / \textbf{56.67} & \textbf{46.67} / 50.00 & \textbf{46.67} / \textbf{60.00} \\
\midrule
\rowcolor{gray!10}
\multicolumn{11}{c}{\textit{Qwen3.5 family (\%)}} \\
\midrule
4B & Base & 66.27 / 79.61 & 58.65 / 70.78 & 5.96 / 11.08 & 74.62 / 82.32 & 53.17 / 67.74 & 87.80 / 94.60 & 33.33 / 43.33 & 36.67 / 53.33 & 36.70 / 50.00 \\
4B & DS-V3.2 Distill & 66.80 / 80.11 & 57.67 / 71.09 & 6.20 / 12.36 & 73.06 / 86.57 & 60.69 / 76.73 & 87.60 / 96.50 & 50.00 / 53.33 & 43.33 / 53.33 & 46.67 / 56.67 \\
4B & MegaScience & 66.97 / 80.57 & 57.36 / 71.87 & 6.32 / 13.04 & 74.70 / 85.23 & 60.00 / 78.90 & 87.00 / 95.40 & 53.33 / 60.00 & 43.33 / 63.33 & 50.00 / 53.33 \\
4B & OpenThought & 68.91 / 79.87 & 58.97 / \textbf{74.18} & 6.84 / 13.28 & 75.02 / 87.90 & 57.66 / 76.30 & \textbf{90.20} / \textbf{96.60} & 50.00 / \textbf{67.78} & 50.00 / \textbf{66.67} & 56.67 / \textbf{66.67} \\
4B & \MYMETHOD & \textbf{69.70} / \textbf{81.11} & \textbf{62.67} / 73.52 & \textbf{7.56} / \textbf{14.04} & \textbf{78.24} / \textbf{88.81} & \textbf{66.18} / \textbf{79.62} & 89.80 / 96.40 & \textbf{56.67} / 60.00 & \textbf{53.33} / 60.00 & \textbf{60.00} / 63.33 \\
\midrule
9B & Base & 69.60 / 80.19 & 63.20 / 77.17 & 6.24 / 12.04 & 78.71 / 90.97 & 51.73 / 69.08 & 89.20 / 96.80 & 40.00 / 46.67 & 40.00 / 60.00 & 40.00 / 60.00 \\
9B & DS-V3.2 Distill & 71.31 / 82.92 & 62.90 / 78.05 & 6.64 / 13.92 & 79.97 / 90.42 & 65.03 / 79.05 & 90.80 / 97.20 & 43.33 / 66.67 & 43.33 / 60.00 & 43.33 / 70.00 \\
9B & MegaScience & 71.52 / 82.10 & 63.53 / 78.48 & 6.36 / 13.52 & 81.30 / 93.64 & 65.75 / 77.75 & 91.20 / 96.20 & 50.00 / 66.67 & 60.00 / 66.67 & 53.33 / 63.33 \\
9B & OpenThought & 73.13 / 82.32 & 64.67 / 77.53 & 7.12 / 13.08 & 77.37 / 92.38 & 67.48 / 80.78 & \textbf{92.00} / 97.60 & 60.00 / 70.00 & \textbf{70.00} / \textbf{80.00} & 60.00 / \textbf{73.33} \\
9B & \MYMETHOD & \textbf{73.96} / \textbf{84.79} & \textbf{67.86} / \textbf{80.21} & \textbf{8.12} / \textbf{14.88} & \textbf{86.80} / \textbf{95.12} & \textbf{71.39} / \textbf{85.11} & 91.80 / \textbf{98.20} & \textbf{63.33} / \textbf{73.33} & \textbf{70.00} / 76.67 & \textbf{66.67} / \textbf{73.33} \\
\bottomrule
\end{tabular}
}
\end{table}

As shown in Table ~\ref{tab:main_results}, the results indicate that \MYMETHOD~consistently outperforms base-model inference and all baselines across the nine benchmarks. For Qwen3-4B, \MYMETHOD~improves pass@3 on every benchmark over the base model, with especially large gains on mathematical reasoning: MATH-500 rises from 90.20 to 98.60, HMMT-Feb.\ from 26.67 to 43.33, HMMT-Nov.\ from 30.00 to 40.00, and AIME 2025 from 33.33 to 53.33. The same pattern appears for Qwen3-8B, where the largest improvements again occur on harder competition-style benchmarks and SciBench.

Compared with DS-V3.2 Distill, which generates reasoning trajectories for the original source problems without structural decomposition, \MYMETHOD~demonstrates that composing new problems from thought modes provides additional benefit beyond trajectory-level distillation. Against MegaScience and OpenThought, two external large-scale reasoning datasets, \MYMETHOD~achieves the best pass@1 on the General-and-STEM benchmarks (CS-Bench, ChemBench, HLE, MedQA, SciBench) for every model size while using a controlled SFT budget of 9,230 examples, suggesting that principled compositional synthesis can be more data-efficient than large-scale external data collection. The Qwen3.5-9B results further confirm that the gains generalize to stronger base models: \MYMETHOD~achieves the best pass@1 on almost all nine benchmarks for this configuration.

\subsection{Ablation Study}
\label{sec:ablation}

We conduct ablation studies to evaluate the contribution of each component of \MYMETHOD. Table~\ref{tab:ablation} presents results when individual components are removed or replaced with simpler alternatives.

\begin{table}[t]
\centering
\caption{Ablation study. Each cell reports pass@1 / pass@3 (\%), with the best value in each block in bold. \textit{w/o Scarcity} sets $\alpha{=}1.0$ so selector sampling uses similarity only. \textit{w/o Filter} skips rollout-based filtering while still applying source-provenance filtering. \textit{Random Retriever} replaces learned retrieval with random candidates. \textit{w/o Reverse Eng.} builds the thought-mode bank directly from rollouts.}
\label{tab:ablation}
\tiny
\renewcommand{\arraystretch}{0.88}
\setlength{\tabcolsep}{2.3pt}
\resizebox{\textwidth}{!}{%
\begin{tabular}{rlccccccccc}
\toprule
\multirow{2}{*}{\textbf{Model}} & \multirow{2}{*}{\textbf{Variant}} & \multicolumn{5}{c}{\textbf{General and STEM}} & \multicolumn{4}{c}{\textbf{Mathematical Reasoning}} \\
\cmidrule(lr){3-7} \cmidrule(lr){8-11}
& & \textbf{CS-Bench} & \textbf{ChemBench} & \textbf{HLE} & \textbf{MedQA} & \textbf{SciBench} & \textbf{MATH-500} & \textbf{HMMT-Feb.} & \textbf{HMMT-Nov.} & \textbf{AIME 2025} \\
\midrule
\rowcolor{gray!10}
\multicolumn{11}{c}{\textit{Qwen3-2507 family (\%)}} \\
\midrule
4B & w/o Scarcity & 65.44 / 78.46 & 57.08 / 71.95 & 5.24 / 11.04 & 59.86 / 78.00 & 51.87 / 66.76 & \textbf{92.40} / 96.80 & \textbf{33.33} / \textbf{43.33} & 26.67 / \textbf{43.33} & 30.00 / 46.67 \\
4B & w/o Filter & 53.45 / 72.92 & 50.16 / 74.59 & 4.28 / 10.84 & 50.20 / 72.82 & 30.63 / 46.24 & 67.80 / 82.00 & 13.33 / 20.00 & 13.33 / 16.67 & 16.67 / 20.00 \\
4B & Random Retriever & 63.04 / 77.68 & 50.28 / 70.77 & 4.89 / 10.44 & 54.20 / 74.78 & 46.89 / 64.40 & 80.20 / 92.42 & 16.00 / 20.00 & 16.67 / 30.00 & 26.67 / 33.33 \\
4B & w/o Reverse Eng. & 61.02 / 77.64 & 50.47 / 71.83 & 4.96 / 11.16 & 57.34 / 75.49 & 45.81 / 63.44 & 81.00 / 91.20 & 23.33 / 36.67 & 23.33 / 30.00 & 33.33 / 40.00 \\
4B & \MYMETHOD & \textbf{68.29} / \textbf{80.16} & \textbf{61.72} / \textbf{74.78} & \textbf{5.96} / \textbf{12.12} & \textbf{65.12} / \textbf{81.30} & \textbf{53.61} / \textbf{71.53} & 85.20 / \textbf{98.60} & \textbf{33.33} / \textbf{43.33} & \textbf{33.33} / 40.00 & \textbf{40.00} / \textbf{53.33} \\
\midrule
8B & w/o Scarcity & 67.92 / 79.99 & 58.10 / 71.75 & 5.96 / 11.20 & 68.42 / 82.33 & 54.77 / 68.50 & \textbf{87.20} / 95.00 & 40.00 / 53.33 & 40.00 / 46.67 & 36.67 / 50.00 \\
8B & w/o Filter & 53.62 / 72.92 & 49.88 / \textbf{73.72} & 5.04 / \textbf{11.76} & 53.50 / 76.36 & 29.19 / 49.85 & 58.40 / 72.40 & 20.00 / 26.67 & 20.00 / 33.33 & 13.33 / 23.33 \\
8B & Random Retriever & 63.74 / 79.08 & 54.29 / 70.61 & 5.68 / 10.92 & 64.57 / 82.64 & 49.86 / 68.79 & 83.40 / 93.00 & 33.33 / 36.67 & 26.67 / 40.00 & 26.67 / 33.33 \\
8B & w/o Reverse Eng. & 64.24 / 79.74 & 54.05 / 70.26 & 5.56 / 10.32 & 63.32 / 82.40 & 49.57 / 68.41 & 83.20 / 93.00 & 26.67 / 43.33 & 24.00 / 40.00 & 30.00 / 36.67 \\
8B & \MYMETHOD & \textbf{69.65} / \textbf{81.40} & \textbf{59.04} / 72.97 & \textbf{6.64} / 11.64 & \textbf{70.77} / \textbf{86.80} & \textbf{57.08} / \textbf{73.41} & 86.80 / \textbf{97.40} & \textbf{43.33} / \textbf{56.67} & \textbf{46.67} / \textbf{50.00} & \textbf{46.67} / \textbf{60.00} \\
\midrule
\rowcolor{gray!10}
\multicolumn{11}{c}{\textit{Qwen3.5 family (\%)}} \\
\midrule
4B & w/o Scarcity & 69.33 / 80.69 & 61.64 / 72.27 & 5.96 / 12.60 & 74.78 / \textbf{89.32} & 63.44 / 75.29 & 89.20 / 95.80 & 36.67 / 50.00 & 46.67 / 50.00 & 43.33 / \textbf{63.33} \\
4B & w/o Filter & 54.90 / 68.91 & 52.68 / 69.00 & 6.12 / 12.16 & 63.63 / 81.62 & 38.29 / 53.61 & 58.60 / 72.60 & 16.67 / 30.00 & 13.33 / 26.67 & 16.67 / 36.67 \\
4B & Random Retriever & 67.01 / 79.62 & 58.10 / 71.13 & 5.44 / 11.68 & 73.61 / 87.27 & 61.99 / 76.73 & 87.40 / 94.20 & 40.00 / 50.00 & 40.00 / 53.33 & 36.67 / \textbf{63.33} \\
4B & w/o Reverse Eng. & 66.96 / 79.37 & 57.31 / 71.56 & 6.48 / 12.92 & 75.02 / 88.69 & 63.01 / 76.45 & 88.80 / 95.40 & 33.33 / 36.67 & 30.00 / 46.67 & 46.67 / 56.67 \\
4B & \MYMETHOD & \textbf{69.70} / \textbf{81.11} & \textbf{62.67} / \textbf{73.52} & \textbf{7.56} / \textbf{14.04} & \textbf{78.24} / 88.81 & \textbf{66.18} / \textbf{79.62} & \textbf{89.80} / \textbf{96.40} & \textbf{56.67} / \textbf{60.00} & \textbf{53.33} / \textbf{60.00} & \textbf{60.00} / 63.33 \\
\midrule
9B & w/o Scarcity & 72.26 / 83.67 & 62.66 / 75.73 & 7.44 / 14.40 & 81.85 / 93.72 & 67.49 / 80.78 & 89.80 / 97.60 & 60.00 / 70.00 & 53.33 / 70.00 & 60.00 / \textbf{76.67} \\
9B & w/o Filter & 60.36 / 73.91 & 55.51 / 73.80 & 5.48 / 12.56 & 70.38 / 85.55 & 41.47 / 56.65 & 58.80 / 81.00 & 23.33 / 30.00 & 26.67 / 30.00 & 26.67 / 43.33 \\
9B & Random Retriever & 70.44 / 81.98 & 59.90 / 74.70 & 6.68 / 13.56 & 80.05 / 91.28 & 65.32 / 78.90 & 90.60 / 97.80 & 56.67 / 63.33 & 46.67 / 60.00 & 56.67 / 70.00 \\
9B & w/o Reverse Eng. & 70.97 / 82.80 & 62.20 / 73.92 & 7.52 / 14.56 & 80.36 / 92.14 & 65.32 / 78.61 & 90.20 / 97.80 & 46.67 / 60.00 & 50.00 / 66.67 & 43.33 / 66.67 \\
9B & \MYMETHOD & \textbf{73.96} / \textbf{84.79} & \textbf{67.86} / \textbf{80.21} & \textbf{8.12} / \textbf{14.88} & \textbf{86.80} / \textbf{95.12} & \textbf{71.39} / \textbf{85.11} & \textbf{91.80} / \textbf{98.20} & \textbf{63.33} / \textbf{73.33} & \textbf{70.00} / \textbf{76.67} & \textbf{66.67} / 73.33 \\
\bottomrule
\end{tabular}
}
\end{table}
Each ablation variant removes or replaces one pipeline component while keeping the rest fixed. The Qwen3-4B result for \textit{w/o Scarcity} shows that removing the scarcity reward from selector scoring weakens most metrics relative to the full \MYMETHOD~ pipeline, including CS-Bench, ChemBench, HLE, MATH-500, and HMMT-Feb. This is consistent with the intended role of the scarcity term: it prevents synthesis from repeatedly selecting common high-similarity thought modes and improves coverage over less frequent reasoning patterns. The remaining ablation variants test complementary hypotheses: \textit{w/o Filter} isolates the value of rollout-based judging before SFT conversion, \textit{Random Retriever} tests whether learned semantic retrieval matters beyond random candidate pools, and \textit{w/o Reverse Eng.} tests whether extracting structured thought modes from solutions is better than building the thought-mode bank directly from raw rollouts. The Qwen3-8B results follow a consistent pattern: the full pipeline achieves the best or near-best scores on all nine benchmarks, and the \textit{w/o Filter} variant again shows the largest degradation, confirming that rollout-based filtering is the single most impactful component across model sizes.

\subsection{Analysis}
\label{sec:analysis}

\begin{figure}[t]
    \centering
    \includegraphics[width=0.95\linewidth]{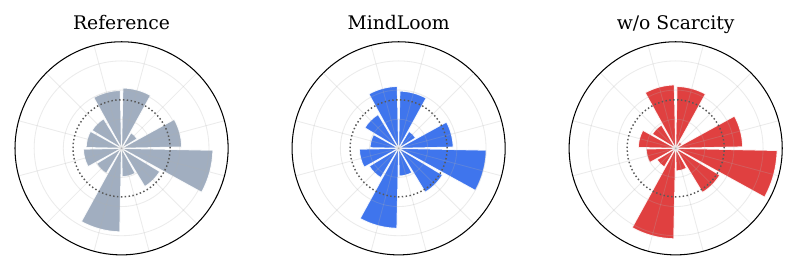}
    \caption{Coverage over the twelve thought-mode clusters (Appendix~\ref{app:taxonomy}). Polar bars give the proportion of selector calls landing in each cluster; the dotted ring marks uniform coverage at $1/12$. \MYMETHOD~tracks the reference distribution; \textit{w/o Scarcity} skews toward popular clusters and starves the smallest one.}
    \label{fig:distribution}
\end{figure}

\textbf{Distribution Analysis.}
We verify that the distribution-aligned selector broadens reasoning-pattern coverage without sacrificing per-instance quality. Figure~\ref{fig:distribution} contrasts the realized distribution $P_\mathrm{gen}$ against the reference $P_\mathrm{ref}$ across the twelve thought-mode clusters defined in Appendix~\ref{app:taxonomy}. \MYMETHOD~tracks $P_\mathrm{ref}$ closely, while \textit{w/o Scarcity} concentrates on a few popular clusters and starves rarer ones; the same scarcity removal also underperforms \MYMETHOD~on hard mathematical reasoning in Table~\ref{tab:ablation}. The two views together indicate that the scarcity term improves benchmark performance \emph{by} broadening reasoning coverage, not at its expense.

\textbf{Difficulty Analysis.}
We characterize the difficulty profile of synthesized questions from rollout outcomes. Roughly half of the synthesized questions are not fully solved by the rollout model under three independent rollouts, and about $15\%$ land in the partial-success zone that supplies the frontier-like SFT records (full counts in Appendix~\ref{app:rollout_filtering_details}). The synthesis pipeline therefore produces a meaningful spread of difficulties rather than uniformly easy or unsolvable problems, validating the rollout-based filtering rationale in Section~\ref{sec:rollout_filtering}.

\textbf{Case Studies.}
We trace one full pipeline run end-to-end to verify that each method stage executes the logic specified in Section~\ref{sec:Method}. The reverse-engineering trace (Appendix~\ref{app:case_re}) walks through the polar-area decomposition: a self-contained seed question is constructed from the last solution window, and one iterative absorption step recovers the full polar-area integral, with the extracted thought mode separating transferable structure from instance-specific bounds. The compositional-synthesis trace (Appendix~\ref{app:case_synthesis}) starts from a risk-free-rate seed and applies three thought modes drawn from three distinct clusters, recovering a competition-level mathematical-finance problem and confirming that the distribution-aligned selector avoids repeated cluster selection. Failure-mode examples (Appendix~\ref{app:failure_cases}) further document the synthesis-incompatibility and reverse-engineering-saturation cases that the pipeline correctly rejects.

\textbf{Hyperparameter Sensitivity.}
Two hyperparameters most directly govern the pipeline: the scarcity weight $\alpha$ that balances semantic compatibility against coverage in distribution-aligned synthesis (Section~\ref{sec:synthesis}), and the reverse-engineering window length $w$ that controls thought-mode granularity (Section~\ref{sec:extraction}). Figure~\ref{fig:sensitivity} (Appendix~\ref{app:hyperparameter_sweep}) sweeps each on Qwen3-4B and Qwen3.5-4B with the other held fixed: performance peaks at the defaults ($\alpha{=}0.65$, $w{=}2$) on both models and varies smoothly within a wide neighborhood, indicating that the operating point is robust rather than finely tuned. Both axes do degrade at the extremes, in directions consistent with the design rationale: very small $\alpha$ removes the compatibility signal and the selector applies thought modes the question state cannot support, while very large $\alpha$ removes scarcity and collapses generation onto popular clusters, recovering the \textit{w/o Scarcity} regime; very small $w$ fragments the bank into hyper-local transformations that lose reusable structure, while very large $w$ bundles multiple reasoning steps into a single mode and erodes the atomicity that retrieval and composition rely on.

%% file: sections/5_Conclusion.tex
\section{Conclusion}
\label{sec:Conclusion}

We have proposed the concept of thought modes, defined as atomic knowledge-reasoning transformations that can be composed to modulate problem complexity. We have presented \MYMETHOD, a framework for synthesizing frontier-level reasoning data through compositional thought mode engineering. \MYMETHOD~ extracts thought mode chains from existing hard problems via reverse engineering, learns to retrieve compatible thought modes through embedding-based matching, composes new problems through distribution-aligned iterative synthesis, and prepares training data through rollout-based judging, source-aware benchmark filtering, and judged-correct SFT conversion. Experiments on nine benchmarks across multiple model families show that models trained on \MYMETHOD-generated data consistently improve over base-model inference, trajectory-level distillation, and external-data baselines, with the strongest gains on competition-style mathematical reasoning. Ablation studies indicate the contribution of each pipeline component, and further analysis points to broad coverage of diverse reasoning types with useful difficulty control.

%% file: sections/Appendix.tex

\appendix

\section{Limitations and Future Work}
The primary limitation of \MYMETHOD~ is that the diversity of synthesized problems is bounded by the thought-mode bank, which is itself derived from the collected reference corpus. If the corpus lacks certain types of advanced reasoning patterns, those patterns will be absent from the bank and therefore from the generated data. Expanding the reference corpus to cover a broader range of domains and difficulty levels is a direct way to alleviate this. Due to computational constraints, we have not yet conducted experiments on larger models or combined \MYMETHOD-generated data with reinforcement learning. Nevertheless, we have validated the approach across two model families and nine benchmarks to provide broad coverage within the available budget. Future work will scale to larger models, integrate with RL-based reasoning training, and extend the pipeline to multi-modal settings.

\section{Thought Mode Format and Examples}
\label{app:thought_mode_format}

This section provides the complete formal specification of the thought mode representation and illustrative examples.

\subsection{Formal Specification}

Each thought mode $\mathcal{T}$ is represented as a four-component tuple $\mathcal{T} = (S_\mathrm{sum}, S_\mathrm{det}, K_\mathrm{gen}, K_\mathrm{spec})$. Table~\ref{tab:thought_mode_fields} describes each component.

\begin{table}[h]
\centering
\caption{Components of a thought mode tuple.}
\label{tab:thought_mode_fields}
\begin{tabular}{lp{9cm}}
\toprule
\textbf{Component} & \textbf{Description} \\
\midrule
$S_\mathrm{sum}$ & A concise, one-sentence summary of the transformation type. It describes what kind of reasoning complexity is added at a high level (e.g., ``Introduce definite integral computation over a bounded domain''). \\
$S_\mathrm{det}$ & A detailed description of the specific modification made to the problem. It specifies exactly how the problem is changed, including which explicit givens are removed and what new derivation is required. \\
$K_\mathrm{gen}$ & The general, domain-transferable knowledge required to handle this thought mode. This includes mathematical theorems, physical laws, algorithmic principles, or other reusable knowledge that applies beyond the specific problem instance (e.g., ``Integration by parts: $\int u\,dv = uv - \int v\,du$''). \\
$K_\mathrm{spec}$ & The problem-specific parameters and constraints involved in this transformation. This includes numerical values, boundary conditions, domain-specific setup, or other details that are unique to the particular problem (e.g., ``Integration bounds: $[0, 2\pi]$; integrand involves $\sin^2(x)$''). \\
\bottomrule
\end{tabular}
\end{table}

The design rationale behind this four-component structure is to separate \textit{what} a transformation does ($S_\mathrm{sum}$, $S_\mathrm{det}$) from \textit{what knowledge} it requires ($K_\mathrm{gen}$, $K_\mathrm{spec}$), and to further separate \textit{transferable} aspects ($S_\mathrm{sum}$, $K_\mathrm{gen}$) from \textit{instance-specific} aspects ($S_\mathrm{det}$, $K_\mathrm{spec}$). This separation enables the retrieval model to match thought modes based on structural similarity (via $S_\mathrm{sum}$ and $K_\mathrm{gen}$) while preserving enough detail (via $S_\mathrm{det}$ and $K_\mathrm{spec}$) for accurate problem synthesis.

\subsection{Extracted Example}

A concrete example of a thought mode tuple is presented in the main text (Section~\ref{sec:analysis}). Here we provide additional context on the tuple format using the same polar-area example. The original problem asks for the area of the polar region $1 \le r \le 2$, $0 \le \theta \le \pi/2$. The reverse-engineering output first creates a seed question that gives the intermediate radial integral $\int_1^2 r\,dr = 3/2$ and asks only for the final outer integration. One extracted thought mode then removes this intermediate result and restores the need to set up and evaluate the polar-area integral. The full JSON output produced by the iterative evolution prompt is:
\begin{verbatim}
{
  "Q_next": "Find the area of the region ...",
  "answer": "3*pi/4",
  "solution_steps": ["Recall area formula ...", "..."],
  "S_sum": "[Knowledge+Reasoning]: Requires recalling
    the polar area formula ...",
  "S_det": "Replace the given radial-integral result
    with the full description of the region ...",
  "K_gen": ["area formula in polar coordinates",
    "setting up double integrals", ...],
  "K_spec": ["radial bounds [1,2]",
    "angular bounds [0, pi/2]"]
}
\end{verbatim}
This output format matches the structured parsing described in Section~\ref{sec:extraction}.

\section{Reverse Engineering Details}
\label{app:reverse_engineering}

\subsection{Forward Inference Configuration}
\label{app:forward_inference}

Before reverse engineering, we run forward inference on the reference corpus to obtain verified solution traces. DeepSeek V3.2 generates three independent rollouts per problem via the OpenAI-compatible API with the following system prompt:
\begin{verbatim}
You are an expert problem solver. Think step by step
and produce a JSON object.
Output format (strict JSON):
{"steps": ["step 1", "step 2", ...],
 "final_answer": "concise final result"}
Rules:
- Each step must be non-trivial and task-specific.
- Avoid macro steps or trivial restatements.
- Keep steps granular but meaningful.
- Do not add extra keys; do not wrap in Markdown.
\end{verbatim}
Each rollout is then judged by the same LLM judge described in Appendix~\ref{app:rollout_filtering_details}. Problems are categorized by rollout outcomes: all-correct, partial (at least one correct), and all-wrong. Only problems with partial-success rollouts are eligible for reverse engineering, as they provide verified solution traces while indicating appropriate difficulty. For all-wrong problems we run a secondary inference pass with DeepSeek V3.2's thinking mode to recover additional verified solution traces; if any rescue rollout is judged correct, the problem becomes eligible and its judged-correct rollout joins the candidate pool consumed by the standard SFT converter (Appendix~\ref{app:sft_format}). The rescue rollouts contribute only to the standard SFT pipeline; no reasoning-mode SFT records are produced for any experiment reported in this paper.

\subsection{Solution Window Partitioning}

Given a verified solution $S = [s_1, s_2, \ldots, s_m]$ with $m$ steps, we partition it into $k$ windows from the tail. Each window contains $w$ consecutive solution steps (the \textit{step size}). The last window $W_k$ covers steps $[s_{m-w+1}, \ldots, s_m]$, the second-to-last window $W_{k-1}$ covers $[s_{m-2w+1}, \ldots, s_{m-w}]$, and so on. If the total number of steps $m$ is not evenly divisible by $w$, the first window $W_1$ contains the remaining $m - (k-1)w$ steps.

The step size $w$ controls the granularity of thought modes. A smaller $w$ yields finer-grained thought modes that capture more atomic reasoning steps, while a larger $w$ produces coarser thought modes that bundle multiple reasoning steps together. We set $w=2$ in our experiments, matching the reverse-engineering launcher used to build the current thought-mode bank (see Section~\ref{app:step_size_analysis}).

\subsection{Seed Generation Prompt}
\label{app:seed_prompt}

The $\operatorname{Seed}$ operation takes the last solution window $W_k$ and generates a minimal seed question $Q_0$. The prompt instructs the LLM to:

\begin{enumerate}[leftmargin=*]
\item Identify all values and quantities that window $W_k$ uses but does not derive (i.e., values computed in earlier windows).
\item Convert each such value into an explicit given in the seed question.
\item Formulate a self-contained question whose complete solution requires exactly the steps in $W_k$.
\item Verify that the seed question is independently solvable without any external information.
\end{enumerate}

The implementation uses the following prompt skeleton. The system message defines the role as a problem designer and reverse-engineering specialist, explains that the model should backtrack from the final solution window, and emphasizes dependency isolation. The user message then provides the subject, original problem, final answer, full solution steps, and the current tail window. The required output is strict JSON:
\begin{verbatim}
{
  "seed_question": "...",
  "answer": "...",
  "solution_steps": ["..."]
}
\end{verbatim}
The prompt explicitly instructs the model to identify upstream-dependent quantities used by the tail window, convert them into explicit givens, and ensure that the seed question can be solved using only the current window.

\subsection{Iterative Evolution Prompt}
\label{app:evolution_prompt}

The $\operatorname{Absorb}$ operation takes the current question $Q_{i-1}$, the next solution window $W_{k-i}$, and the remaining earlier solution steps. The prompt instructs the LLM to:

\begin{enumerate}[leftmargin=*]
\item Identify which explicit given in $Q_{i-1}$ is the result of the computation in window $W_{k-i}$.
\item Remove that given from the problem statement and modify the question so that the solver must derive the value.
\item Ensure the evolved question $Q_i$ remains well-defined and solvable.
\item Extract the thought mode tuple $\mathcal{T}_i = (S_\mathrm{sum}, S_\mathrm{det}, K_\mathrm{gen}, K_\mathrm{spec})$ that describes the added reasoning requirement.
\end{enumerate}

The iterative evolution prompt receives the original target problem, the current intermediate question, the upstream solution steps, and the next solution window to absorb. It asks the model to remove one explicit dependency from the current question and rewrite the problem so that the solver must derive it. The output is strict JSON with both the evolved question and the thought-mode tuple:
\begin{verbatim}
{
  "Q_next": "...",
  "answer": "...",
  "solution_steps": ["..."],
  "S_sum": "...",
  "S_det": "...",
  "K_gen": ["..."],
  "K_spec": ["..."]
}
\end{verbatim}
The parser enforces these fields and validates that all string-list fields are lists of strings before the step is added to the chain.

\subsection{Quality Control}

To ensure the quality of extracted thought mode chains, we apply several automatic checks:

\begin{itemize}[leftmargin=*]
\item \textbf{Candidate selection.} Reverse engineering is run only on source questions with useful successful rollouts: non-thinking partial-success records are selected directly, and all-wrong non-thinking records are rescued only when the corresponding thinking-mode rollout has at least one judged-correct solution. All-correct, incomplete, and unrescued all-wrong records are skipped.
\item \textbf{Strict structured parsing.} Seed outputs must contain \texttt{seed\_question}, \texttt{answer}, and \texttt{solution\_steps}; iterative outputs must contain \texttt{Q\_next}, \texttt{answer}, \texttt{solution\_steps}, \texttt{S\_sum}, \texttt{S\_det}, \texttt{K\_gen}, and \texttt{K\_spec}. The parser rejects invalid JSON, missing keys, and non-string-list knowledge fields.
\item \textbf{Resume-safe deduplication.} Existing reverse-engineered outputs are keyed by question text or id and skipped on resumed runs, preventing duplicate chains from entering the retained file.
\end{itemize}

The retained reverse-engineering file contains 8,322 records, with 32,269 extracted thought-mode steps in total and an average chain length of 3.88. These records are selected from the 58,526 source problems through strict eligibility and quality criteria. Only problems with partial-success rollouts---indicating appropriate difficulty and providing verified solution traces---are candidates for extraction; all-correct problems are excluded as too easy for meaningful thought mode decomposition, and problems without any correct rollout lack the verified traces needed for reverse analysis. Among eligible candidates, the structured parser enforces valid JSON and complete thought mode tuples, further ensuring extraction quality.

\subsection{Step Size Analysis}
\label{app:step_size_analysis}

The reverse-engineering window length $w$ controls thought-mode granularity. We use $w=2$ as the default: a smaller value creates very local transformations and inflates extraction cost, while a larger value merges multiple dependencies into a coarser, less reusable thought mode. Empirical sweeps over $w\in\{1,2,3,4\}$ are reported jointly with the scarcity-weight sweep in Appendix~\ref{app:hyperparameter_sweep}.

\section{Retrieval Model Details}
\label{app:retrieval_details}

\subsection{Text Serialization Format}

The problem state and thought mode are serialized into text for the embedding model as follows.

\textbf{Retriever query encoding.}
The retriever uses the raw intermediate question text as the query-side input without a \texttt{[STATE]} prefix.

\textbf{Selector state encoding.}
The distribution-aware selector embeds the current generation state as:
\begin{verbatim}
[STATE] Question: {question} || Answer: {answer}
        || Solution: {solution_steps_joined_by_bar}
\end{verbatim}

\textbf{Thought mode encoding:}
\begin{verbatim}
[LOGIC] Summary: {S_sum} || Detail: {S_det}
        || K_gen: {K_gen} || K_spec: {K_spec}
\end{verbatim}
where each field corresponds to one component of the thought mode tuple $\mathcal{T}$.

The \texttt{[STATE]} and \texttt{[LOGIC]} prefixes serve as type indicators in selector embeddings, while the fine-tuned retriever uses raw question text on the query side and the \texttt{[LOGIC]} representation on the thought-mode side. The delimiter \texttt{||} separates individual fields.

\subsection{Model Architecture}

The retrieval model is based on Qwen3-Embedding-0.6B, a pre-trained text embedding model. We use it as a bi-encoder: one encoder processes the problem state, and the same encoder processes the thought mode. Both inputs share the same encoder weights. The implementation supports optional L2 normalization, but we do not enable it in our experiments; FAISS therefore uses inner-product search over raw retriever embeddings. The selector's separate general-purpose embedding model computes cosine similarity explicitly.

\subsection{ANCE Hard Negative Mining Procedure}

We follow an ANCE-style protocol~\citep{xiong2021approximate} for asynchronous hard negative mining:

\begin{enumerate}[leftmargin=*]
\item \textbf{Initialization.} At the start of training, we build a FAISS~\citep{douze2026faiss} inner-product index using the initial model embeddings over all thought modes in the bank $\mathcal{B}$.
\item \textbf{Mining.} For each training pair $(Q_i, \mathcal{T}_i^+)$, we query the FAISS index with the embedding of $Q_i$ and retrieve the top-$k$ most similar thought modes, excluding the positive $\mathcal{T}_i^+$. These serve as hard negatives $\{\mathcal{T}_j^-\}$.
\item \textbf{Refresh.} After every $R$ training steps, we re-encode all thought modes using the updated model and rebuild the FAISS index. This ensures the hard negatives remain informative as training progresses.
\end{enumerate}

\subsection{Training Hyperparameters}

\begin{table}[h]
\centering
\caption{Retrieval model training hyperparameters.}
\label{tab:retrieval_hyperparams}
\begin{tabular}{ll}
\toprule
\textbf{Hyperparameter} & \textbf{Value} \\
\midrule
Base model & Qwen3-Embedding-0.6B \\
Learning rate & $2 \times 10^{-5}$ \\
Batch size & 2 \\
Gradient accumulation & 1 \\
Training epochs & 10, capped at 300 optimizer steps \\
Maximum optimizer steps & 300 \\
Hard negatives per sample ($k$) & 5 \\
Margin ($\gamma$) & 0.2 \\
FAISS refresh interval ($R$) & 5 steps \\
Optimizer & AdamW \\
Weight decay & 0.01 \\
Warmup steps & 100 \\
Max sequence length & 512 \\
Precision / distributed training & bf16 with DeepSpeed ZeRO-2 \\
Embedding normalization & Disabled in the main launcher \\
Train/validation split & 80/20 by source id \\
Validation / checkpoint cadence & Evaluate every step; save every 10 steps \\
Checkpoint used by generation & step 20 in the main launcher \\
\bottomrule
\end{tabular}
\end{table}

The checkpoint at step~20 is selected based on validation-set performance during training. Training runs for up to 300 optimizer steps with evaluation at every step; the step~20 checkpoint represents an early point where the model has learned meaningful compatibility signals without overfitting to the training pairs. Later checkpoints showed diminishing returns on the 20\% held-out validation split.

\section{Synthesis and Rollout Filtering Details}
\label{app:synthesis_details}

\subsection{Compositional Synthesis Algorithm}

Algorithm~\ref{alg:synthesis} presents the full synthesis procedure.

\begin{algorithm}[h]
\caption{Distribution-Aligned Compositional Synthesis}
\label{alg:synthesis}
\begin{algorithmic}[1]
\REQUIRE Seed question $Q_0$, thought mode bank $\mathcal{B}$, retrieval model $f$, number of evolution steps $n$, clusters $\{c_1, \ldots, c_K\}$, reference distribution $P_\mathrm{ref}$
\ENSURE Evolved question $Q_n$, applied thought modes $[\mathcal{T}_1, \ldots, \mathcal{T}_n]$
\STATE Initialize generated-count vector from cache if available, otherwise zeros
\FOR{$i = 0$ to $n-1$}
    \STATE Encode problem state: $\mathbf{e}_q \leftarrow f(Q_i)$
    \STATE Retrieve top-$m$ candidate thought modes: $\{\mathcal{T}_{j_1}, \ldots, \mathcal{T}_{j_m}\} \leftarrow \operatorname{Retrieve}(\mathbf{e}_q, \mathcal{B}, m)$
    \STATE Compute selection scores via Equation~\ref{eq:score}
    \STATE Sample $\mathcal{T}_{i+1}$ via softmax sampling (Equation~\ref{eq:softmax})
    \STATE $Q_{i+1}, \texttt{is\_compatible} \leftarrow \operatorname{Apply}(Q_i, \mathcal{T}_{i+1})$ \COMMENT{LLM-based transformation}
    \IF{\texttt{is\_compatible} is false and $i=0$}
        \STATE Record a seed/original fallback output for the first step
    \ELSIF{\texttt{is\_compatible} is false}
        \STATE Terminate synthesis for this seed
    \ENDIF
    \STATE Increment generated count for the selected cluster and persist cache
\ENDFOR
\RETURN $Q_n$, $[\mathcal{T}_1, \ldots, \mathcal{T}_n]$
\end{algorithmic}
\end{algorithm}

\subsection{Compatibility Verification}

During synthesis, each thought mode application is verified for compatibility. The LLM-based $\operatorname{Apply}$ function checks the following criteria:

\begin{itemize}[leftmargin=*]
\item \textbf{Coherence.} The evolved question $Q_{i+1}$ must be a well-formed, grammatically correct question.
\item \textbf{Solvability.} The evolved question must have a definite, derivable answer.
\item \textbf{Relevance.} The thought mode $\mathcal{T}_{i+1}$ must meaningfully increase the difficulty of the question, not merely rephrase it.
\item \textbf{Consistency.} The evolved question must not contradict any information from the original seed or previously applied thought modes.
\end{itemize}

If any criterion is violated, the application is rejected. An incompatible first step falls back to the seed/original bookkeeping state, while an incompatible later step terminates synthesis for that seed. The generated record stores \texttt{is\_compatible}, the selected \texttt{logic\_id}, model provenance, and the evolved state for each accepted step. The generation stage produces 7,515 candidate records before later rollout filtering.

Across 12,636 total synthesis steps, 8,614 (68.2\%) were judged compatible. The remaining 4,022 (31.8\%) were first-step incompatibilities that triggered seed fallback. No later-step terminations occurred in the main run, indicating that once a seed successfully accepts its first thought mode, subsequent steps maintain coherence.

\subsection{Synthesis Prompt Template}
\label{app:synthesis_prompt}

The $\operatorname{Apply}$ function uses a structured LLM prompt to fuse a selected thought mode with the current problem state. The system message defines the role as a problem designer and provides seven macroscopic difficulty enhancement strategies: (A)~target upgrade (from single-point to structural or global quantities), (B)~parameterization and generalization (from concrete values to general forms), (C)~constraint coupling (introducing interdependent conditions that form simultaneous or boundary constraints), (D)~reasoning chain extension (multi-stage dependencies where later steps require earlier results), (E)~inverse or constructive formulations (deriving conditions from results or constructing objects under constraints), (F)~uncertainty and robustness analysis (error bounds, worst-case analysis, sensitivity), and (G)~algorithmic and resource constraints (requiring implementable solutions with complexity bounds).

The prompt enforces five correctness criteria: \textit{solvability} (the evolved question must have a derivable solution), \textit{uniqueness} (the answer must be determinate or disambiguated by a stated rule), \textit{condition consistency} (new constraints must not contradict existing ones), \textit{condition necessity} (every added condition must be explicitly used in the solution), and \textit{closed-loop coherence} (the question, conditions, and solution must form a complete reasoning chain).

The user message provides the current problem state (question, answer, solution steps) and the selected thought mode ($S_\mathrm{sum}$, $S_\mathrm{det}$, $K_\mathrm{gen}$, $K_\mathrm{spec}$). The required output is strict JSON:
\begin{verbatim}
{
  "is_compatible": true/false,
  "question": "evolved problem statement",
  "answer": "final answer",
  "solution_steps": ["step 1", "step 2", ...],
  "S_sum": "logic summary",
  "S_det": "detailed modification description",
  "K_gen": ["general knowledge 1", ...],
  "K_spec": ["specific parameter 1", ...]
}
\end{verbatim}
The \texttt{is\_compatible} flag allows the model to reject incompatible pairings, which triggers the fallback logic described in Algorithm~\ref{alg:synthesis}.

\subsection{Synthesis Hyperparameters}

\begin{table}[h]
\centering
\caption{Compositional synthesis hyperparameters.}
\label{tab:synthesis_hyperparams}
\begin{tabular}{ll}
\toprule
\textbf{Hyperparameter} & \textbf{Value} \\
\midrule
Chat model & DeepSeek V3.2 \\
Selector embedding model & Qwen3-Embedding-0.6B \\
Reverse engineering step size $w$ & 2 \\
Evolution steps $n$ & 3 \\
Retrieval candidate pool & top-20 \\
K-Means clusters $K$ & 12 \\
Similarity weight $\alpha$ & 0.65 \\
Selector temperature $\tau$ & 1.0 \\
Scarcity smoothing $\epsilon$ & $10^{-3}$ \\
Cluster minimum similarity & 0.0 \\
\bottomrule
\end{tabular}
\end{table}

\subsection{SFT and Evaluation Hyperparameters}

\begin{table}[h]
\centering
\caption{Fine-tuning and evaluation hyperparameters.}
\label{tab:sft_hyperparams}
\begin{tabular}{ll}
\toprule
\textbf{Hyperparameter} & \textbf{Value} \\
\midrule
\multicolumn{2}{l}{\textit{Fine-tuning}} \\
Framework & ms-swift, full-parameter SFT \\
Precision & bf16 \\
Maximum sequence length & 4096 \\
Per-device batch size & 4 \\
Gradient accumulation steps & 8 \\
Learning rate & $3 \times 10^{-5}$ \\
Warmup ratio & 0.05 \\
Training epochs & 3 \\
\midrule
\multicolumn{2}{l}{\textit{Evaluation}} \\
Inference engine & vLLM \\
Rollouts per test item & 3 \\
Temperature & 0.7 \\
Maximum generation length & 8096 \\
Maximum model length & 23000 \\
Judging & API-based LLM judge \\
\bottomrule
\end{tabular}
\end{table}

\subsection{Data Pipeline Statistics and Domain Composition}
\label{app:data_composition}

This section traces how data flows through each pipeline stage and reports the domain composition at each step. Table~\ref{tab:pipeline_flow} summarizes the pipeline stages and record counts.

\begin{table}[h]
\centering
\caption{Data pipeline flow and record counts at each stage.}
\label{tab:pipeline_flow}
\small
\begin{tabular}{lr}
\toprule
\textbf{Stage} & \textbf{Records} \\
\midrule
Reference corpus & 58,526 \\
Forward inference rollouts & 58,526 \\
Reverse engineering & 8,322 \\
Compositional synthesis & 7,515 seeds $\rightarrow$ 11,433 rollout records \\
Source-provenance filtering & 10,854 \\
Final SFT conversion & 9,230 \\
\bottomrule
\end{tabular}
\end{table}

\textbf{Seed question sources.}
Seed questions are taken from the reverse-engineering output. For each original problem with a valid reverse-engineered chain, the pipeline stores a self-contained seed question, answer, and solution steps. All 8,322 reverse-engineered records are used as candidate seeds. During synthesis, previously generated outputs are deduplicated by question ID, so resumed runs skip already-processed seeds. Of the 8,322 candidates, 7,515 produced at least one accepted synthesis step (including seed-fallback outputs from incompatible first steps); the remaining 807 were skipped due to deduplication from prior partial runs or processing failures. The 8,322 reverse-engineered records are drawn from 14 source datasets, with SuperGPQA contributing 65.8\%, MMLU-Pro 10.4\%, Hendrycks-MATH 7.5\%, and the remaining 11 datasets collectively 16.3\%.

\textbf{Domain composition of the final SFT set.}
Table~\ref{tab:sft_composition} reports the domain breakdown of the 9,230 \MYMETHOD~ SFT examples. The composition reflects the reference corpus distribution after provenance-based filtering removes records originating from held-out benchmark sources.

\begin{table}[h]
\centering
\caption{Domain composition of the final 9,230 \MYMETHOD~ SFT examples by original seed source.}
\label{tab:sft_composition}
\small
\begin{tabular}{lrr}
\toprule
\textbf{Seed source} & \textbf{Count} & \textbf{\%} \\
\midrule
SuperGPQA & 6,378 & 69.1 \\
MMLU-Pro & 916 & 9.9 \\
Hendrycks-MATH & 876 & 9.5 \\
ARC-AGI-2 & 461 & 5.0 \\
OlympiadBench & 247 & 2.7 \\
TheoremQA & 181 & 2.0 \\
GPQA Diamond & 71 & 0.8 \\
AGIEval & 53 & 0.6 \\
MiniF2F & 33 & 0.4 \\
AMC 2023 & 10 & 0.1 \\
AIME 2024 & 4 & 0.0 \\
\midrule
\textbf{Total} & \textbf{9,230} & \textbf{100.0} \\
\bottomrule
\end{tabular}
\end{table}

The three provenance-filtered sources, HLE, SciBench, and AIME 2025, contributed 388, 188, and 3 records respectively to the pre-filtering pool of 11,433 records. Their removal yields a 94.9\% retention rate.

\textbf{Distillation baseline composition.}
The DS-V3.2 Distill baseline is constructed by prompting DeepSeek V3.2 with the original source problems using the same inference system prompt and rollout configuration as the forward inference stage (Appendix~\ref{app:forward_inference}). The conversion step selects the first judged-correct rollout for each problem to form an SFT record in the same ms-swift format as \MYMETHOD~ records (Appendix~\ref{app:sft_format}). This produces a pool of 40,937 distillation records. This pool is randomly subsampled to match the \MYMETHOD~ budget. The domain composition of the full distillation pool is: SuperGPQA 43.9\%, Hendrycks-MATH 24.2\%, MMLU-Pro 21.5\%, OlympiadBench 5.2\%, and the remaining eight sources collectively 5.2\%. Compared with the \MYMETHOD~ SFT set (Table~\ref{tab:sft_composition}), the distillation pool is more evenly spread across the three largest sources rather than concentrated on SuperGPQA.

\subsection{Rollout Filtering Details}
\label{app:rollout_filtering_details}

\textbf{Rollout Configuration.}
For each generated question $Q$, we perform three independent rollouts in the filtering and evaluation scripts. Filtering rollouts use the OpenAI-compatible provider configured in \texttt{.env}; evaluation rollouts use local vLLM inference with temperature 0.7 by default. The main evaluation launcher sets maximum new tokens to 8096, maximum model length to 23000, tensor parallel size to 2 when enough GPUs are visible, and uses batch size 16 for standard datasets and 4 for HLE. The pass rate is computed as the fraction of non-empty parsed rollouts that produce a correct final answer.

\textbf{Automated Judging.}
We use DeepSeek V3.2 as the LLM-based judge to determine answer correctness. The judge receives the generated question $Q$, the reference answer stored with the generated record, the rationale or solution steps, and the model's response. The system message is:
\begin{verbatim}
You are a strict grader. For each problem, you receive
the question, the authoritative ground-truth answer,
and a short rationale explaining why that answer is
correct. You also receive a model's predicted answer.
Decide whether the predicted answer should be counted
as correct based on the question and the rationale.
Return JSON: {"is_correct": true/false}.

Guidance:
- Judge semantic correctness, not exact string match.
- Use the rationale to understand what counts as a
  valid answer for this question.
- If the predicted answer misses the required outcome,
  mark false.
- Do not add extra keys or text.
\end{verbatim}
The user message concatenates the question, ground-truth answer, rationale, and model answer, followed by \texttt{``Respond ONLY with the JSON object.''}. The same prompt is shared by filtering and evaluation.

\textbf{Discrete Difficulty Labels and Conversion.}
The implemented filter uses a discrete three-rollout criterion rather than a continuous threshold sweep. Questions for which all rollouts are correct are labeled all-correct; questions with one or two correct rollouts are labeled partial; and questions with zero correct rollouts are labeled all-wrong. The hard-item export retains partial and all-wrong questions for analysis. SFT conversion is stricter in a different way: it writes a training example only if a judged-correct response is available, using the non-thinking rollout first and the paired thinking rollout as fallback.

The generator produced 7,515 candidate records and 12,636 accepted or recorded step-level outputs. The non-thinking rollout file contains 11,433 generated rollout records, of which 11,427 are complete; after held-out-source filtering, 10,854 records remain. The reasoning rollout file contains 11,331 records, and the hard-item export yields 4,106 records from the partial and all-wrong categories. The final SFT conversion writes 9,230 non-thinking records and 9,067 thinking-format records. To keep SFT comparisons controlled by data quality rather than data quantity, the DS-V3.2 distillation pool is subsampled to match the \MYMETHOD~ budget.

\subsection{SFT Training Record Format}
\label{app:sft_format}

Each SFT training record follows the ms-swift conversation format with three messages. The user message embeds the inference system prompt followed by the generated question, and the assistant message contains the first judged-correct rollout response:
\begin{verbatim}
{"messages": [
  {"role": "system",
   "content": "You are a helpful assistant"},
  {"role": "user",
   "content": "You are an expert problem solver. Think
    step by step and respond with a strict JSON object.
    Output format:
    {\"steps\": [...], \"final_answer\": \"...\"}
    Rules: [step quality rules as in inference]
    Problem: {question}"},
  {"role": "assistant",
   "content": "{\"steps\": [...],
    \"final_answer\": \"...\"}"}
]}
\end{verbatim}

\section{Reference Corpus and Evaluation Benchmarks}
\label{app:corpus_details}

\subsection{Source Datasets}

The thought-mode reference bank is constructed from 58,526 source problems aggregated from 16 datasets across three source files. Table~\ref{tab:source_datasets} lists each dataset, its approximate size, and its domain coverage.

\begin{table}[h]
\centering
\caption{Source datasets used to construct the thought-mode reference bank.}
\label{tab:source_datasets}
\small
\begin{tabular}{llr}
\toprule
\textbf{Dataset} & \textbf{Domain} & \textbf{Approx.\ size} \\
\midrule
SuperGPQA~\citep{pteam2025supergpqascalingllmevaluation} & Multi-domain expert QA & 26,529 \\
MMLU-Pro~\citep{wang2024mmlupro} & Multi-domain multiple choice & 12,102 \\
Hendrycks-MATH~\citep{hendrycks2021measuringmathematicalproblemsolving} & Competition mathematics & 10,114 \\
OlympiadBench~\citep{he2024olympiadbenchchallengingbenchmarkpromoting} & Olympiad mathematics and science & 2,665 \\
HLE~\citep{phan2026hle} & Expert-level reasoning & 2,500 \\
ARC-AGI-2~\citep{chollet2025arcagi2} & Abstraction and reasoning & 1,120 \\
AGIEval~\citep{zhong2024agieval} & Standardized exam problems & 1,000 \\
TheoremQA~\citep{chen2023theoremqa} & Theorem-based reasoning & 731 \\
SciBench~\citep{wang2024scibenchevaluatingcollegelevelscientific} & College-level science & 692 \\
MiniF2F~\citep{zheng2021minif2f} & Formal mathematical proofs & 392 \\
ProofNet~\citep{azerbayev2023proofnet} & Formal proofs & 371 \\
GPQA Diamond~\citep{rein2023gpqagraduatelevelgoogleproofqa} & Graduate-level science QA & 198 \\
AMC 2023~\citep{zwhe99amc23} & AMC competition problems & 40 \\
AIME 2024~\citep{jia2025aime2024} & AIME competition problems & 30 \\
AIME 2025~\citep{balunovic2025matharena} & AIME competition problems & 30 \\
MathArena Apex~\citep{balunovic2025matharena} & Competition mathematics & 12 \\
\midrule
\textbf{Total} & & \textbf{58,526} \\
\bottomrule
\end{tabular}
\end{table}

\subsection{Contamination Control}

Before SFT conversion, the pipeline traces each synthesized item back to its seed-source provenance and removes generated examples whose origin matches a held-out benchmark source ID. Table~\ref{tab:contamination_filter} summarizes the filtering configuration.

\begin{table}[h]
\centering
\caption{Provenance-based contamination filters applied before SFT conversion.}
\label{tab:contamination_filter}
\small
\begin{tabular}{lll}
\toprule
\textbf{Filtered source ID} & \textbf{Reason} & \textbf{Overlapping benchmark} \\
\midrule
\texttt{opencompass\_\_AIME2025} & Direct overlap & AIME 2025 \\
\texttt{xw27\_\_scibench} & Direct overlap & SciBench \\
\texttt{cais\_\_hle} & Direct overlap & HLE \\
\bottomrule
\end{tabular}
\end{table}

The Hendrycks-MATH records in our reference corpus are drawn from the training split of the dataset; we additionally remove the 500 questions that constitute the MATH-500 evaluation set to prevent any leakage. Because the remaining Hendrycks-MATH records have no overlap with MATH-500, they are retained in the SFT set. Similarly, the AGIEval records in our corpus do not overlap with any evaluation benchmark. The remaining evaluation benchmarks, CS-Bench, ChemBench, MedQA, HMMT February 2025, and HMMT November 2025, have no overlap with any source dataset in the reference corpus and therefore require no source-level filtering.

During evaluation, we skip all test items that require image inputs, as the current pipeline operates in a text-only setting. The reported test-item counts in Table~\ref{tab:eval_benchmarks} reflect the text-only subsets actually evaluated.

\subsection{Evaluation Benchmarks}

Table~\ref{tab:eval_benchmarks} summarizes the nine evaluation benchmarks. We report pass@1 and pass@3 for each benchmark, defined as the fraction of test items for which one or three independent rollouts produce at least one correct answer, respectively.

\begin{table}[h]
\centering
\caption{Evaluation benchmarks.}
\label{tab:eval_benchmarks}
\small
\begin{tabular}{llr}
\toprule
\textbf{Benchmark} & \textbf{Domain} & \textbf{Test items} \\
\midrule
CS-Bench~\citep{song2025csbenchcomprehensivebenchmarklarge} & Computer science & 2,419 \\
ChemBench~\citep{mirza2024largelanguagemodelssuperhuman} & Chemistry & 2,542 \\
MATH-500~\citep{hendrycks2021measuringmathematicalproblemsolving} & Competition mathematics & 500 \\
HLE~\citep{phan2026hle} & Expert-level reasoning & 2,500 \\
HMMT Feb.\ 2025~\citep{balunovic2025matharena} & Competition mathematics & 30 \\
HMMT Nov.\ 2025~\citep{balunovic2025matharena} & Competition mathematics & 30 \\
MedQA~\citep{jin2020diseasedoespatienthave} & Medicine & 1,273 \\
AIME 2025~\citep{balunovic2025matharena} & Competition mathematics & 30 \\
SciBench~\citep{wang2024scibenchevaluatingcollegelevelscientific} & College-level science & 692 \\
\bottomrule
\end{tabular}
\end{table}

\section{Extended Case Studies}
\label{app:case_studies}

This appendix gives concrete walkthroughs of the four-stage pipeline. Reverse-engineering examples (Appendix~\ref{app:case_re}) illustrate Section~\ref{sec:extraction}; compositional-synthesis examples (Appendix~\ref{app:case_synthesis}) illustrate Section~\ref{sec:synthesis}; failure-case analysis (Appendix~\ref{app:failure_cases}) documents the two main failure modes that the rollout-based filtering stage is designed to catch.

\subsection{Reverse Engineering: Worked Examples}
\label{app:case_re}

\textbf{Example R1: Polar Area.}
The original problem asks for the area of the polar region $1 \le r \le 2$, $0 \le \theta \le \pi/2$. Reverse engineering first creates a seed question that gives the intermediate radial integral $\int_1^2 r\,dr = 3/2$ as an explicit given and asks only for the final outer integration. This seed is self-contained and solvable in a single step (Phase~1 of the extraction process). The subsequent iterative evolution step removes the radial-integral result and requires the solver to set up the full polar-area double integral from the region bounds. The extracted thought mode for this step is:
\begin{itemize}[leftmargin=*,itemsep=1pt]
\item $S_\mathrm{sum}$: ``Requires recalling the polar area formula, setting up the correct double integral with the given bounds, and performing a sequence of basic calculus computations.''
\item $S_\mathrm{det}$: ``Replace the given radial-integral result with the full description of the region in polar coordinates and require the solver to compute the integral from the bounds.''
\item $K_\mathrm{gen}$: area formula in polar coordinates, setting up double integrals, evaluating definite integrals.
\item $K_\mathrm{spec}$: radial bounds $[1, 2]$, angular bounds $[0, \pi/2]$.
\end{itemize}
This example shows the separation between transferable reasoning structure ($S_\mathrm{sum}$, $K_\mathrm{gen}$) and instance-specific details ($S_\mathrm{det}$, $K_\mathrm{spec}$). The $K_\mathrm{gen}$ component (``polar area formula, double integrals'') can guide retrieval for other calculus problems, while $K_\mathrm{spec}$ records the concrete numerical setup.

\textbf{Example R2: Combinatorial Probability.}
The original problem asks: ``Bob and Alice each have a bag that contains one ball of each of the colors, blue, green, orange, red, and violet. Alice randomly selects one ball from her bag and puts it into Bob's bag. Bob then randomly selects one ball from his bag and puts it into Alice's bag. What is the probability that after this process the contents of the two bags are the same?'' The answer is $1/3$.

\textbf{Seed ($Q_0$).} The seed question provides the intermediate state explicitly: after Alice gives Bob a ball of color $X$, Bob's bag contains 6 balls, of which exactly 2 are color $X$. The question asks only for the probability that Bob picks a ball of color $X$ from his 6 balls. The solution is $2/6 = 1/3$ in a single step.

\textbf{Step~1.} The seed is evolved to ask the full restoration question: what is the probability that the final bag contents match the initial state? The solver must now deduce that Bob must return the duplicate color and then compute the conditional probability $2/6$. The extracted thought mode is:
\begin{itemize}[leftmargin=*,itemsep=1pt]
\item $S_\mathrm{sum}$: ``Requires analyzing the necessary condition for the final bag states to be identical and performing the conditional probability calculation within the symmetry framework.''
\item $K_\mathrm{gen}$: conditional probability, set reasoning for state-matching, symmetry arguments.
\item $K_\mathrm{spec}$: five colors, two-step transfer process, bags must return to initial composition.
\end{itemize}

\textbf{Step~2.} The necessary condition (``Bob must return color $X$'') is removed from the problem statement; the solver must derive it by analyzing the state transitions. $K_\mathrm{gen}$ adds state transition analysis in combinatorial processes.

\textbf{Step~3.} The intermediate state description (``Bob's bag has 6 balls with a duplicate'') is removed; only the initial identical-bags condition and the transfer process remain. The solver must deduce the intermediate state from first principles.

\textbf{Step~4.} The final evolved question recovers the original problem: no intermediate state, no condition hints. The full chain length is 4, with each step removing one layer of explicit scaffolding. The progression illustrates how reverse engineering decomposes a problem by successively re-introducing dependencies that the seed had converted into explicit givens.

\subsection{Compositional Synthesis: Worked Examples}
\label{app:case_synthesis}

Each synthesis record stores the seed, up to three applied steps, selected logic IDs, thought-mode metadata, model provenance, compatibility flags, and the evolved question/answer/solution state at each step. We walk through one three-step synthesis trace in detail.

\textbf{No-Arbitrage / Risk-Free Rate Derivation.}
The seed question is a risk-free rate computation problem from mathematical finance. It provides a pre-derived equation $-0.1\lambda = -0.05$ and asks to solve for $\lambda$, then substitute into the no-arbitrage condition to find $r = 0.02$. Three thought modes are then applied, each drawn from a different cluster of the taxonomy in Table~\ref{tab:taxonomy}.

\textbf{Step~1 (Cluster~5, Algebraic Manipulation).} The selected thought mode originates from the same problem's reverse-engineering chain and has $S_\mathrm{sum}$: ``The user must now understand the derivation of the key linear equation by performing the subtraction of the two no-arbitrage conditions.'' After application, the evolved question removes the pre-derived equation and requires the solver to eliminate $r$ from two given no-arbitrage equations. Compatibility: \texttt{true}.

\textbf{Step~2 (Cluster~3, Physics \& Engineering).} The selected thought mode originates from a photoresistor calibration problem and has $S_\mathrm{sum}$: ``Absorption of the explicit equation-solving step for the linear model. The solver must now set up and algebraically solve the system of two linear equations.'' After application, the numerical equations are also removed; the solver must substitute $(\mu_i, \sigma_i)$ into the general relationship before solving. Compatibility: \texttt{true}.

\textbf{Step~3 (Cluster~4, Applied Quantitative).} The selected thought mode requires theoretical derivation: ``The risk-neutral drift condition is now required to be applied explicitly, rather than being implicitly used.'' The final question asks the solver to derive the no-arbitrage relationship from first principles using It\^{o}'s lemma and Girsanov's theorem, set up the system, and solve. Compatibility: \texttt{true}.

The three steps draw from three distinct clusters (5, 3, 4), illustrating how the distribution-aligned selector avoids repeated cluster selections. The final question is substantially harder than the seed: it requires stochastic calculus foundations in addition to algebraic manipulation.

\subsection{Failure Case Analysis}
\label{app:failure_cases}

The pipeline exposes two main failure modes. Reverse engineering can fail when the model output is invalid JSON or omits required tuple fields; such records are rejected by the parser. Synthesis can fail when the selected thought mode is incompatible with the current question state; the generation parser requires an explicit \texttt{is\_compatible} flag, and incompatible later steps terminate the chain rather than adding an ill-formed question to the output.

\textbf{Synthesis incompatibility.} A right-triangle cosine seed question (``In right triangle $DEF$ with right angle at $D$, $DE=24$, $EF=25$. What is $\cos(E)$?'') was paired with a thought mode from a tangent-simplification problem whose $S_\mathrm{sum}$ requires ``combining fraction arithmetic and the simplification of a complex fraction with nested radicals.'' The $K_\mathrm{gen}$ of the applied mode includes sum-of-tangent identities, product-to-sum conversions, and denominator rationalization. These are fundamentally incompatible with the seed: a basic adjacent-over-hypotenuse calculation has no intermediate expressions to rationalize and no trigonometric sums to simplify. The model correctly set \texttt{is\_compatible=false} and the pipeline recorded a seed-fallback output for this step.

\textbf{Reverse engineering saturation.} In a chain extracted from the absolute-value optimization problem ``Let $f(x) = |x-p| + |x-15| + |x-p-15|$, where $0<p<15$. Determine the minimum value of $f(x)$ for $x \in [p, 15]$,'' the reverse engineering process produced a 4-step chain. However, at step~3 the model returned the same question text as step~2, with empty $K_\mathrm{gen}$ and $K_\mathrm{spec}$ fields. The $S_\mathrm{det}$ explanation noted that ``the interpretation step was already implicitly included in the reasoning chain of the current question.'' This represents a saturation failure: the problem's solution structure had been fully decomposed by earlier steps, and further absorption could not produce a genuinely simpler intermediate question. The empty knowledge fields signal this degeneracy to downstream processing.

\section{Thought Mode Taxonomy and Statistics}
\label{app:taxonomy}

The distribution-aligned selector uses $K{=}12$ K-Means clusters over thought-mode embeddings. Table~\ref{tab:taxonomy} provides a taxonomy of these clusters based on the $S_\mathrm{sum}$ fields of the thought modes assigned to each cluster. Labels are derived by inspecting representative thought modes near each cluster center.

\begin{table}[h]
\centering
\caption{Thought mode cluster taxonomy. Counts reflect the total number of times each cluster was selected during the main synthesis run (43,384 total selections). The three representative $S_\mathrm{sum}$ excerpts for each cluster are drawn from thought modes closest to the cluster center.}
\label{tab:taxonomy}
\small
\renewcommand{\arraystretch}{1.05}
\begin{tabular}{clrrl}
\toprule
\textbf{ID} & \textbf{Label} & \textbf{Count} & \textbf{\%} & \textbf{Representative $S_\mathrm{sum}$ excerpt} \\
\midrule
0 & Factual \& Historical Recall & 4,483 & 10.3 & Recall a biographical fact; chronological deduction \\
1 & Geometric \& Vector Reasoning & 1,361 & 3.1 & Dihedral angle volume formula; Law of Sines \\
2 & Biomedical \& Clinical & 4,495 & 10.4 & Receptor blockade mechanism; culture media \\
3 & Physics \& Engineering & 4,312 & 9.9 & Wedge product; PSD first-null derivation \\
4 & Applied Quantitative & 2,575 & 5.9 & Duration calculation with rounding; convergence \\
5 & Algebraic Manipulation & 2,642 & 6.1 & Expand and simplify; Hessian determinant \\
6 & Grid Pattern Recognition & 2,844 & 6.6 & Coordinate-based diamond augmentation; rotation \\
7 & Discrete Math \& Structures & 2,186 & 5.0 & Combinatorial counting under fold constraints \\
8 & Life Sciences \& Interdisciplinary & 6,224 & 14.3 & Gene epistasis; codon mutation reasoning \\
9 & Number Theory \& Counting & 2,120 & 4.9 & Legendre's formula; palindrome enumeration \\
10 & Physical Chemistry & 3,338 & 7.7 & Bohr energy derivation; van der Waals relation \\
11 & Social Sciences \& Conceptual & 6,804 & 15.7 & JTB epistemology; economic model elimination \\
\midrule
& \textbf{Total} & \textbf{43,384} & \textbf{100} & \\
\bottomrule
\end{tabular}
\end{table}

The two largest clusters are Life Sciences (Cluster~8, 14.3\%) and Social Sciences (Cluster~11, 15.7\%), which together account for 30.0\% of all selections. The smallest cluster is Geometric \& Vector Reasoning (Cluster~1, 3.1\%). This imbalance reflects the domain composition of the reference corpus, where SuperGPQA (which spans social sciences, humanities, and life sciences) contributes the majority of source problems (Table~\ref{tab:sft_composition}). The scarcity reward in the distribution-aligned selector compensates for this imbalance by boosting underrepresented clusters during synthesis: despite Cluster~1 containing only 3.1\% of selections, it still received 1,361 selections rather than being starved.

The 12 clusters span a broad taxonomy: STEM reasoning (Clusters~1, 3, 5, 9, 10), biomedical and life sciences (Clusters~2, 8), spatial and pattern reasoning (Cluster~6), applied quantitative reasoning (Cluster~4), discrete mathematics (Cluster~7), factual recall (Cluster~0), and social sciences (Cluster~11). Cluster~6 is distinctive in that it is dominated by grid-transformation tasks (primarily from ARC-AGI-2), where the thought modes describe coordinate-based pattern extraction and rule validation rather than mathematical derivation.

\begin{figure}[h]
    \centering
    \includegraphics[width=\linewidth]{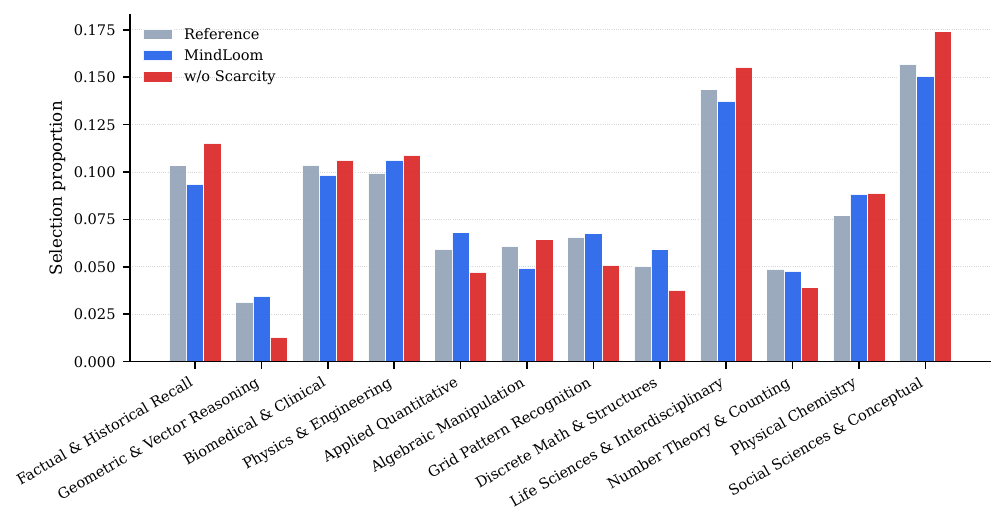}
    \caption{Per-cluster selection proportions for the three selector variants. Cluster labels follow Table~\ref{tab:taxonomy}.}
    \label{fig:distribution_full}
\end{figure}

Figure~\ref{fig:distribution_full} extends the body view (Figure~\ref{fig:distribution}) with the full per-cluster proportions for the three selector variants discussed in Section~\ref{sec:analysis}. \MYMETHOD~stays close to $P_\mathrm{ref}$ across all twelve clusters, whereas \textit{w/o Scarcity} systematically over-samples the most common clusters and underrepresents rare ones such as Geometric \& Vector Reasoning (Cluster~1).

\section{Hyperparameter Sensitivity Sweep}
\label{app:hyperparameter_sweep}

Two hyperparameters most directly govern the pipeline: the scarcity weight $\alpha$ in distribution-aligned synthesis (Section~\ref{sec:synthesis}), which mediates the balance--quality tradeoff, and the reverse-engineering window length $w$ (Section~\ref{sec:extraction}; see also Appendix~\ref{app:step_size_analysis}), which controls thought-mode granularity. We sweep each in turn on two foundation models (Qwen3-4B and Qwen3.5-4B) while holding the other at its default, and report the average pass@3 over the nine benchmarks. The $\alpha{=}1.0$ row in the $\alpha$ sweep coincides with the \textit{w/o Scarcity} ablation in Table~\ref{tab:ablation}; the default operating point ($\alpha{=}0.65$, $w{=}2$) coincides with the \MYMETHOD~row in Table~\ref{tab:main_results}.

\begin{figure}[h]
    \centering
    \includegraphics[width=\linewidth]{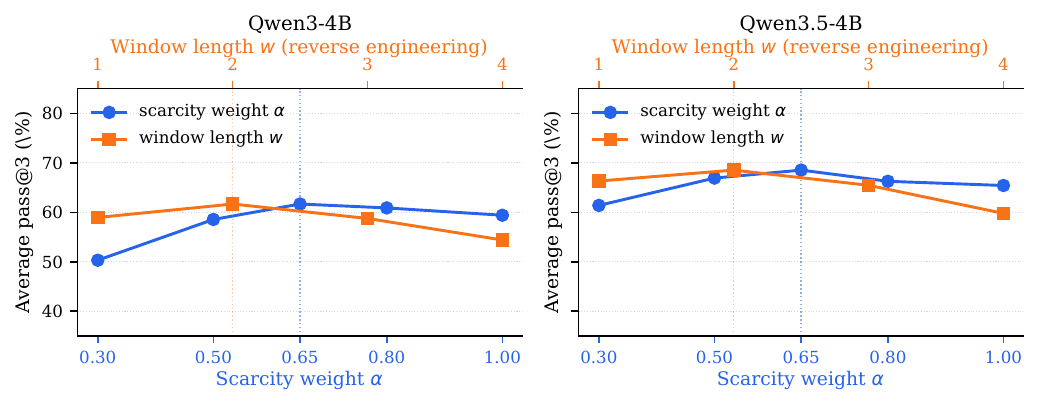}
    \caption{Hyperparameter sensitivity sweeps on Qwen3-4B (left) and Qwen3.5-4B (right). Bottom axis: scarcity weight $\alpha$ (default $0.65$). Top axis: reverse-engineering window length $w$ (default $2$). Vertical y-axis: average pass@3 across the nine benchmarks. Dotted vertical lines mark the defaults.}
    \label{fig:sensitivity}
\end{figure}

The two-parameter scope reflects the dominant axes of pipeline behavior: $\alpha$ governs the selector's exploration--exploitation tradeoff (already isolated in Table~\ref{tab:ablation}'s scarcity ablation), and $w$ governs the granularity of the upstream thought-mode bank itself. Other selector knobs ($K$, $n$, $\tau$) primarily affect surface-level synthesis behavior rather than the qualitative shape of the data distribution.